%% file: main_arxiv.tex
\documentclass[acmtog,nonacm]{acmart}

\input{acm_shared}

\setcopyright{none}
\renewcommand\footnotetextcopyrightpermission[1]{}

\makeatletter
\global\@ACM@balancefalse
\makeatother

\begin{document}
\input{acm_frontmatter}

\makeatletter
\let\hyxmp@parse@acmart\relax
\renewcommand{\@mktitle}{\@mktitle@iii}
\renewcommand{\@mkauthors}{%
  \global\setbox\mktitle@bx=\vbox{%
    \noindent\unvbox\mktitle@bx
    \centering\large
    Sagi Polaczek\textsuperscript{1}\qquad
    Noa Kraicer\textsuperscript{1}\qquad
    Gal Metzer\textsuperscript{1}\qquad
    Zhuo Ning\textsuperscript{2}\\
    Ali Mahdavi-Amiri\textsuperscript{2}\qquad
    Daniel Cohen-Or\textsuperscript{1}\qquad
    Raja Giryes\textsuperscript{1}\\[0.35em]
    {\small
      \textsuperscript{1}Tel Aviv University\qquad
      \textsuperscript{2}Simon Fraser University}
    \par\medskip
  }%
}
\makeatother

\fancypagestyle{standardpagestyle}{%
  \fancyhf{}
  \fancyfoot[C]{\small\thepage}
  \renewcommand{\headrulewidth}{0pt}
  \renewcommand{\footrulewidth}{0pt}
}
\fancypagestyle{firstpagestyle}{%
  \fancyhf{}
  \renewcommand{\headrulewidth}{0pt}
  \renewcommand{\footrulewidth}{0pt}
}
\pagestyle{standardpagestyle}

\authorsaddresses{}

\maketitle
\hypersetup{%
  pdfcreator={LaTeX},
  pdfkeywords={audio-video generation, cross-modal attention,
    diffusion models, inference-time steering, semantic leakage,
    source attribution}
}

\input{sec/1_introduction}
\input{sec/2_related_work}
\input{sec/3_problem_setup}
\input{sec/4_method}
\input{sec/5_experiments}
\input{sec/6_Discussion}
\input{sec/7_acknowledgements}

\bibliographystyle{ACM-Reference-Format}
\bibliography{bib}

\appendix
\input{sec/X_supp}

\end{document}

%% file: acm_shared.tex
\AtBeginDocument{%
  }

\acmJournal{TOG}
\usepackage{amsmath}
\usepackage{amsfonts}
\usepackage{nicefrac}
\usepackage{pgfplots}
\usetikzlibrary{arrows.meta}
\usepgfplotslibrary{groupplots}
\pgfplotsset{compat=1.18}
\usepackage{subcaption}
\usepackage{wrapfig}

\newcommand{\TOGTeaserMode}{}

%% file: acm_frontmatter.tex
\makeatletter
\ifnum\@abspage@last=1\relax
  \xdef\@abspage@last{\number\maxdimen}
  \gdef\PreviousTotalPages{0}
\fi
\makeatother

\title{The Attention Triangle in Audio-Video Models}

\author{Sagi Polaczek}
\affiliation{%
  \institution{Tel Aviv University}
  \city{Tel Aviv}
  \country{Israel}}

\author{Noa Kraicer}
\affiliation{%
  \institution{Tel Aviv University}
  \city{Tel Aviv}
  \country{Israel}}

\author{Gal Metzer}
\affiliation{%
  \institution{Tel Aviv University}
  \city{Tel Aviv}
  \country{Israel}}

\author{Zhuo Ning}
\affiliation{%
  \institution{Simon Fraser University}
  \city{Burnaby}
  \country{Canada}}

\author{Ali Mahdavi-Amiri}
\affiliation{%
  \institution{Simon Fraser University}
  \city{Burnaby}
  \country{Canada}}

\author{Daniel Cohen-Or}
\affiliation{%
  \institution{Tel Aviv University}
  \city{Tel Aviv}
  \country{Israel}}

\author{Raja Giryes}
\affiliation{%
  \institution{Tel Aviv University}
  \city{Tel Aviv}
  \country{Israel}}

\renewcommand{\shortauthors}{Polaczek et al.}

\input{sec/0_abstract}

\keywords{audio-video generation, cross-modal attention, diffusion models,
  inference-time steering, semantic leakage, source attribution}

\input{figures/teaser}

%% file: sec/0_abstract.tex
\begin{abstract}
Audio-video diffusion models rely on cross-modal attention to coordinate text, sound, and visual content, yet this same mechanism can introduce subtle and systematic semantic leakage.
We study these models by probing and analyzing the ``attention triangle,'' comprising the three cross-attention edges connecting the text, audio, and video streams, and examine how semantic information is routed across modalities during generation.
Our analysis reveals that routing along the audio-video edge is bidirectional: audio can influence video generation, while video can influence audio generation.
This edge is shaped by biases encoded in the model's parameters and emerges as a major contributor to leakage: when prompts are in tension with learned priors, cross-modal interactions may override the intended conditioning and reroute semantics toward visually canonical but incorrect outcomes.
These effects suggest that semantic artifacts arise not merely from attention spreading beyond its intended target, but from structured, bias-driven interactions along specific pathways.
Building on this perspective, we extract attention-derived signals that expose how semantics are distributed and grounded across modalities, and use them as a diagnostic tool to both analyze and deliberately incur leakage under controlled conditions.
This enables us to probe the internal dynamics of cross-modal routing and isolate the role of individual interactions. We further leverage these signals to guide inference-time interventions that encourage more consistent cross-modal alignment.
Extensive experiments support our analysis and demonstrate improved semantic grounding while preserving generation quality.
\end{abstract}

%% file: figures/teaser.tex
\newcommand{\attentiontriangleteasercontent}[1]{%
  \centering
  \includegraphics[width=#1]{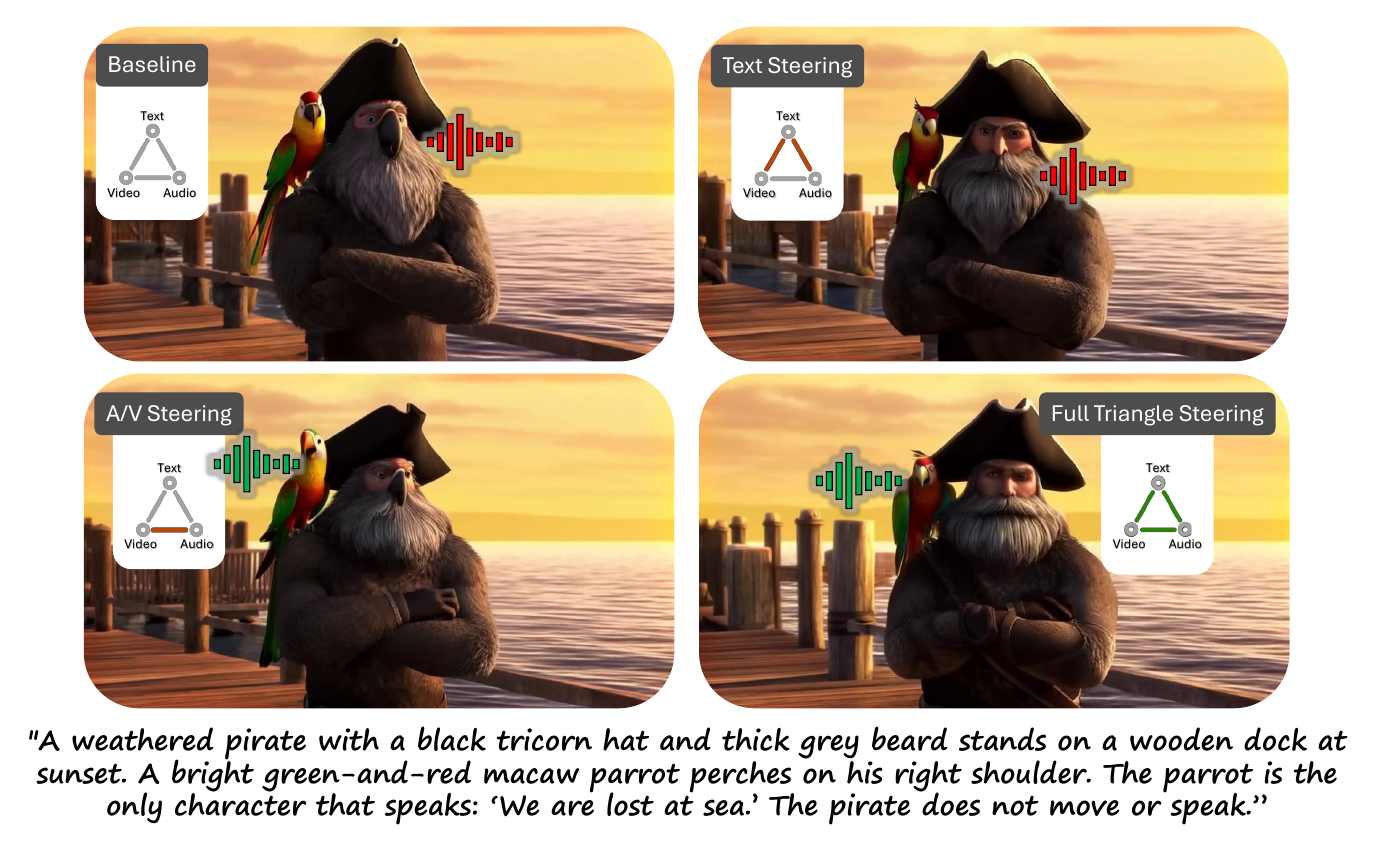}
  \caption{
\textbf{The Attention Triangle in Audio-Video Models.}
\textbf{Top-left:} baseline T2AV generation exhibits strong leakage; the pirate visually inherits parrot-like attributes and becomes the speaking source.
\textbf{Top-right:} Ours-Text restores the pirate's appearance, but speech remains incorrectly localized to the pirate.
\textbf{Bottom-left:} Ours-AV correctly localizes speech to the parrot, but appearance leakage persists in the pirate.
\textbf{Bottom-right:} Ours-Full jointly restores correct appearance and localizes speech to the intended source.
The prompt is shown below the figure.
}
  \Description{Four generated-video examples arranged in a two-by-two grid for a dockside scene with a pirate and a parrot. The baseline gives the pirate parrot-like visual attributes and makes the pirate speak. Ours-Text restores the pirate's appearance but leaves speech on the pirate. Ours-AV moves speech to the parrot but leaves the pirate's visual corruption. Ours-Full preserves the pirate's appearance and assigns speech to the parrot.}
  \label{fig:teaser}
}

\ifdefined\TOGTeaserMode
\begin{teaserfigure}
  \attentiontriangleteasercontent{\textwidth}
\end{teaserfigure}
\else
\begin{figure}[h]
  \attentiontriangleteasercontent{0.8\linewidth}
\end{figure}
\fi

%% file: sec/1_introduction.tex
\ifdefined\TOGTeaserMode\else
  \input{figures/teaser}
\fi

\section{Introduction}
\label{sec:intro}

Recent advances in \emph{diffusion models}, particularly with Diffusion Transformers (DiT) \citep{peebles2023dit}, have established attention as a central mechanism for conditional generation.
By leveraging cross-attention within transformer architectures, these models can associate conditioning signals such as text, reference images, or other modalities with the evolving visual representation.
Attention captures and propagates semantic associations across interacting representations throughout denoising, enabling expressive and compositional generation.
The same mechanism, however, mixes information softly and globally without explicit locality or exclusivity constraints.
Semantic associations can therefore spread beyond their intended targets, producing attribute leakage, semantic entanglement, and unstable bindings over time.

These limitations become more pronounced in audio-video generation, where attention must jointly mediate interactions between text, audio, and video modalities (Figs.~\ref{fig:teaser} and~\ref{fig:t2v_vs_t2av}).
We refer to this trimodal setting as an \emph{attention triangle} (Fig.~\ref{fig:overview}), in which each modality both influences and constrains the others. Unlike bimodal conditioning, alignment is no longer pairwise but coupled across all three modalities. The model must reconcile semantic intent from text, temporal structure from audio, and spatiotemporal realization in video. These coupled constraints can produce competing signals, modality dominance, and entangled associations, allowing a competing interpretation to propagate through attention and reduce global consistency.
Our ablations identify the audio-video pathway as a major contributor to semantic leakage in LTX-2 (Fig.~\ref{fig:steering_comparisons_knight} and Supp.~Fig.~\ref{fig:steering_comparisons_monster}). We hypothesize that this vulnerability arises partly because audio tokens encode temporal (but not spatial) position, leaving source localization underconstrained.

In this work, we revisit audio-video generation through an analysis-driven lens. We analyze the leakage problem, as introduced above, through the structure of the attention triangle, while placing particular emphasis on failures of source attribution, namely whether sound semantics are grounded in the appropriate visual entity. Viewing the model through this lens, leakage can be understood in terms of how semantic information is routed across the text, audio, and video streams. More generally, this routing is inherently bidirectional: audio can influence the visual interpretation, while visual priors can in turn bias the generation and attribution of sound.

An interesting observation that emerges from our analysis is that the audio-video edge of this triangle often acts as a weak link (Fig.~\ref{fig:exp-attn-leakage-v2a2v} and Supp.~Fig.~\ref{fig:supp-ltx-attention-routing}).
In cases where the prompt is in tension with the model's learned cross-modal biases, this interaction may override the binding induced by the text, rerouting sound semantics toward visually canonical but incorrect sources. This suggests that leakage is not only a consequence of semantic associations extending to unintended entities, but is closely tied to bias-driven interactions encoded in the model's parameters, which manifest along specific cross-modal pathways.

Building on this perspective, we extract attention-derived signals that expose source attribution, and use them as a diagnostic tool to both analyze and deliberately incur leakage. By inducing controlled failures, we probe how sound semantics are routed across entities and modalities, and isolate the role of specific cross-modal interactions. We then use these signals to guide inference-time interventions toward more consistent sound-to-source grounding.

A representative example is shown in Fig.~\ref{fig:teaser}. The baseline exhibits both appearance leakage and incorrect source attribution, with the pirate inheriting parrot-like attributes and becoming the speaking source. Partial interventions reveal a decoupling: steering the text edges restores appearance but not attribution, while steering the audio-video edge corrects attribution but not appearance. Only joint steering of all triangle edges resolves both, illustrating that no single interaction suffices to control cross-modal routing.

Across diverse prompts and settings, our experiments reveal consistent leakage patterns and show that the framework supports both probing and mitigation (Tab.~\ref{tab:main_leakage_comparison} and Supp.~Fig.~\ref{fig:user-study-pairwise}).

%% file: sec/2_related_work.tex
\section{Background and Related Work}
\label{sec:related}

Diffusion models rely on two complementary attention mechanisms. \emph{Self-attention} operates within a single modality, allowing each token to aggregate long-range context from all others~\citep{vaswani2017attention}. \emph{Cross-attention} couples two token sequences, typically the noisy latent and a text embedding, so latent tokens selectively attend to relevant conditioning tokens~\citep{rombach2022ldm}. In text-to-image DiT~\citep{peebles2023dit} and UNet backbones, cross-attention serves as the primary mechanism for grounding prompt semantics spatially in the image.

Our work builds on three closely related lines of research: the use of cross-attention as a control surface for image generation, its failure mode of leakage between distinct entities, and audio-video diffusion models that inherit and compound this problem along a new modality axis.

\subsection{Attention Manipulation and Leakage in Image Diffusion Models}
\label{sec:rw:image}

Cross-attention has become the primary control surface of text-to-image diffusion: it encodes spatial layout~\citep{hertz2022prompttoprompt} and carries attribute binding through its keys and values~\citep{feng2023structurediffusion}. Inference-time methods steer it to boost under-attended tokens and recover missing or under-represented subjects~\citep{chefer2023attendandexcite}, align attention maps with syntactic structure to correct attribute correspondence~\citep{rassin2024syngen}, separate competing entities while binding attributes~\citep{li2023divideandbind}, or enforce user-specified layouts~\citep{chen2024trainingfreelayout}. Even in modern RoPE-based MMDiT backbones \citep{esser2024sd3}, different transformer layers play qualitatively distinct roles that invite layer-targeted edits~\citep{wei2025freeflux}. Together these works treat cross-attention as a steerable, semantically meaningful signal yielding fine-grained control without retraining.

The flip side is \emph{leakage}: because attention is a soft, global mixing process without explicit locality or exclusivity constraints, features routinely bleed between entities the prompt intends to keep distinct. A prompt such as ``a striped cat next to a spotted dog'' frequently produces a striped dog or a spotted cat, because patterns are routed by the semantics of the target entity rather than the target description. This failure has been attributed to attention-layer feature blending across visually similar subjects~\citep{dahary2024beyourself}, conflict between externally imposed layouts and the noise prior~\citep{dahary2025bedecisive}, entangled End-of-Sequence text embeddings that aggregate attributes across the prompt~\citep{mun2025ale}, and mis-allocated attention logits across entities~\citep{ventura2025deleaker}; the corresponding remedies all operate as inference-time interventions, treating leakage as an attention-routing failure. A related failure mode is \emph{contextual contradiction}, where one concept implicitly suppresses another through entangled learned associations; \citet{huberman2025contradictory} address this via stage-aware prompting that decomposes the prompt across denoising stages rather than editing attention directly. Our work extends this perspective beyond text-to-image generation to the \textit{attention triangle}, where leakage occurs not only between visual entities but also across modalities.

\subsection{Video and Audio-Video Diffusion Models}
\label{sec:rw:av}

Early text-to-video systems such as Make-A-Video~\citep{singer2022makeavideo} and Imagen Video~\citep{ho2022imagenvideo} attached spatiotemporal modules \citep{ho2022videodiffusion,blattmann2023alignlatents} to pretrained text-to-image models, generating \emph{silent} clips with audio outside the generative loop. The shift to Diffusion Transformer (DiT) backbones~\citep{openai2024sora} enabled large-scale joint video training behind open systems including CogVideoX~\citep{yang2024cogvideox}, HunyuanVideo~\citep{kong2024hunyuanvideo}, Wan~\citep{wan2025wan}, and LTX-Video~\citep{hacohen2024ltxvideo}, alongside closed-source Sora~\citep{openai2024sora} and Veo~\citep{google2025veo}. These systems retain the silent-video assumption.

A more recent thread closes the audio gap by jointly generating audio and video in a single diffusion framework. MM-Diffusion~\citep{ruan2023mmdiffusion} pioneered this by pairing video and audio sub-nets through a random-shift cross-attention block exchanging time-aligned information; MM-LDM~\citep{sun2024mmldm} lifted the idea into a shared semantic latent space. Transformer backbones then produced dual-branch designs coupling two parallel DiT towers via cross-modal attention: SyncFlow~\citep{liu2025syncflow} fuses a dual diffusion-transformer for temporally aligned text-to-audio-video; AV-Link~\citep{hajiali2025avlink} repurposes intermediate features of frozen audio and video diffusion models as cross-modal conditioning; JavisDiT~\citep{liu2025javisdit} introduces a hierarchical spatiotemporal prior aligning the two streams at multiple granularities; and Ovi~\citep{low2025ovi} adopts a fully symmetric twin-DiT design with paired cross-attention layers. On the closed side, Veo~3~\citep{google2025veo} performs joint denoising over a unified token sequence covering both modalities. The open foundation model LTX-2~\citep{hacohen2026ltx2efficientjointaudiovisual} is representative of the state of the art.

Recent work more directly examines or structures cross-modal interaction in dual-stream generators. UniAVGen introduces modality-specific, temporally aligned interaction and learned face-aware modulation~\citep{zhang2026uniavgen}, while Cross-Modal Context Learning (CCL) identifies background-region attention bias, optimization instability, and conflicts between text and cross-modal conditions, addressing them through temporal partitioning, learnable context tokens, and learned routing and guidance~\citep{ma2026ccl}. Very recent work uses bidirectional audio-video attention to guide training-free sparsification and cache reuse while preserving synchronization~\citep{gao2026synchrony}. These approaches redesign or accelerate cross-modal interaction; we instead target entity-level semantic leakage and source attribution in a pretrained generator through training-free interventions across all three triangle edges.

Complementary work evaluates spatial audio-video alignment using visual-object and stereo sound-direction estimates~\citep{shimada2026savgbench}, or introduces trained reference-conditioned branches for multimodal control~\citep{li2026mmcontrol}; our method instead targets prompt-specified source binding without auxiliary training or reference inputs.
  
While these architectures make joint audio-video generation tractable, their cross-modal attention inherits and arguably amplifies the same leakage and source-attribution failures seen in the image domain.
Unlike image leakage between spatially colocated features, the audio-video case introduces a fundamental dimensional mismatch: a 1D audio temporal embedding must attend to a 3D spatiotemporal positional embedding~\citep{hacohen2026ltx2efficientjointaudiovisual}.
Forcing video patches to collapse onto the 1D time axis strips intra-frame spatial distinctness, creating a degeneracy that we diagnose and address in the following sections.

%% file: sec/3_problem_setup.tex
\section{Identifying the Triangle Problem}
\label{sec:problem_setup}
\input{figures/overview}

We consider joint text-to-video and text-to-audio generation with modern video diffusion models that synthesize a video clip and its accompanying soundtrack from a single textual prompt.
Given a textual prompt, the models simultaneously condition both a video stream of frames and a temporally aligned audio stream, and the two modalities are coupled through cross-attention so that what is heard reinforces what is seen and vice versa.
This mutual dependence suggests a useful abstraction: rather than viewing the different conditioning pathways independently, we regard them as a coupled trimodal system in which semantic information can be routed both directly and indirectly between text, audio, and video.
\input{figures/t2v_vs_t2av}

We formalize this system as an \emph{attention triangle} comprising three cross-attention edges (Fig.~\ref{fig:overview}) and associate each surface with a pre-softmax bias matrix. Bias arrows denote conditioning flow: $\mathbf B_{X\to Y}$ acts on the $Y$-query/$X$-key surface. Thus, $\mathbf B_{T\to V}$ and $\mathbf B_{T\to A}$ bias the text-conditioned video and audio surfaces (Eqs.~\ref{eq:t2v_bias} and~\ref{eq:t2a_bias}), while $\mathbf G_{VA}$ denotes the audio-video agreement matrix and $\mathbf B_{A\leftrightarrow V}$ denotes the corresponding pair of audio-video biases (Eqs.~\ref{eq:agreement} and~\ref{eq:agreement_bias}).
While this joint formulation is appealing, it makes the prompt highly susceptible to semantic leakage: textual attributes intended for one entity are silently rerouted to another, more frequent or more salient entity in the scene.
For example, a prompt such as \textit{``a parrot sitting on the shoulder of a pirate, the \emph{parrot} is \emph{talking}''} reliably produces a speaking pirate rather than a talking parrot, because speech carries a strong learned prior toward human-like figures, causing the model to route audio tokens to pirate patches rather than to the parrot the prompt explicitly designates as the speaker.

\subsection{The Attention Triangle}
The attention triangle connects three token populations through pairwise cross-attention: (1) text tokens, (2) audio tokens, and (3) video patches.
Every pair of modalities exchanges information through cross-attention, so a single attribute embedded in the text can reach the video stream both directly (text $\to$ video) and indirectly through the audio stream (text $\to$ audio $\to$ video), and symmetrically in the opposite direction. This relationship is depicted in Figure~\ref{fig:overview}.
Composing these pairwise paths also exposes second-order, within-modality interactions; in particular, the video$\to$audio$\to$video rollout reveals an effective audio-mediated coupling between visual regions, formalized in Eq.~\ref{eq:vis2vis} and visualized in Fig.~\ref{fig:exp-attn-leakage-v2a2v}.
A particularly problematic edge of this triangle is the audio-video cross-attention itself.
The model's training distribution encodes strong priors directly in this edge: speech-like audio features are correlated with human-shaped patches, not parrots. Even when the text correctly assigns speech to the parrot, the audio-video edge can override that binding and drag the sound toward its statistically familiar source.
Beyond these learned statistical biases, we hypothesize that the audio-video edge may also be underconstrained spatially.
In LTX-2~\citep{hacohen2026ltx2efficientjointaudiovisual}, video patches and audio tokens share temporal positional information, while audio tokens do not carry explicit within-frame spatial coordinates.
This asymmetry may force cross-modal correspondence to rely more heavily on learned semantic compatibility between audio content and visual entities than on explicit spatial alignment.
It may therefore make it more difficult for cross-modal attention to distinguish between spatially distinct candidate sources and may contribute to weaker sound localization.

These considerations motivate us to test whether audio-video cross-attention is a major contributor to semantic leakage in LTX-2.
We examine this question through attention visualizations, controlled attention interventions that induce or mitigate leakage, and complementary ablations.
Unlike the direct text-conditioning paths, this edge relays semantic information between two generated modalities along an implicit pathway that is not directly specified by the user's prompt.

\subsection{Audio Leakage Analysis}
\label{sub:attention_vis}

\paragraph{Direction convention.}
We index attention matrices by query and key modality: $\mathbf P_{VA}$ has video queries and audio keys. Under our key/value information-flow convention, this is the $A\to V$ conditioning surface because audio values contribute to video-query outputs; under row rollout, right multiplication by $\mathbf P_{VA}$ maps a video-indexed distribution to an audio-indexed distribution and therefore constitutes a $V\to A$ rollout step.

\paragraph{Visualizing Audio-Mediated Attention}
As a first-order probe, we inspect $\mathbf P_{VA}$, the video-query/audio-key attention matrix.
For each video query, we sum its attention weights over the selected speech-active audio keys and average the resulting spatial score across heads and denoising steps. On leakage prompts, high scores concentrate spatially on the wrong subject; for example, video queries on the pirate assign more attention to speech-active audio keys than video queries on the parrot, even though the prompt assigns speech to the parrot.
This misbinding is visualized in Fig.~\ref{fig:exp-attn-leakage-simple}.

To make matrix orientation explicit, let $\mathbf P_{QK}^{(\ell)}\in[0,1]^{N_Q\times N_K}$ denote the head-averaged attention matrix at layer $\ell$, with rows indexed by queries from modality $Q$ and columns by keys from modality $K$. Following the Markov-chain interpretation of attention matrices~\citep{erel2025attention}, we compose consecutive video-query/audio-key and audio-query/video-key matrices:
\begin{equation}
\mathbf P_{VV}^{\mathrm{eff}}
\;=\;
\mathbf P_{VA}^{(\ell)}\mathbf P_{AV}^{(\ell+1)}
\;\in\;[0,1]^{N_V\times N_V}.
\label{eq:vis2vis}
\end{equation}
The effective matrix $\mathbf P_{VV}^{\mathrm{eff}}$ captures a video-to-video coupling mediated by the audio stream that does not appear in any single attention layer.
Seeding a uniform row distribution over each entity's visual mask and propagating it through $\mathbf P_{VV}^{\mathrm{eff}}$ (see Appendix~\ref{sec:supp-leakage-vis}), we observe that on leakage prompts, the mass migrates from the intended source to the visually canonical one (e.g.\ from the parrot patches to the pirate), making the audio-mediated mis-binding visible as a concrete spatial transport pattern.

\paragraph{Removing the Audio}

A complementary way to test the same hypothesis is to ablate the audio branch and check whether the leakage persists.
LTX-2 ships in audio-free and joint variants sharing the same backbone (see Appendix~\ref{sec:supp-setup}), making this ablation straightforward.
In the paired examples shown in Fig.~\ref{fig:t2v_vs_t2av}, failures present under joint T2AV generation are absent from the audio-free T2V outputs despite the shared visual backbone. Together with the text-edge-only intervention in Sec.~\ref{sec:comparisons}, which leaves source misattribution unresolved, this provides complementary evidence that the audio-video pathway is a major contributor to leakage in LTX-2. These interventions do not, however, establish temporal-only positional encoding as the pathway's unique failure mechanism.

%% file: figures/overview.tex
\begin{figure}[t]
  \centering
  \resizebox{\linewidth}{!}{\input{figures/overview_diagram}}
  \caption{
  \textbf{The Attention Triangle.}
  Edge-specific biases promote attention to the intended speaker (green) and suppress leakage to the competing entity (red).
  \textit{Intended} denotes amplified attention for prompt-designated pairings: \mbox{source$\leftrightarrow$source}, \mbox{sound$\leftrightarrow$sound}, or \mbox{source$\leftrightarrow$sound}. \textit{Conflicting} denotes attenuated pairings between these signals and the competing source.
  The agreement matrix $\mathbf G_{VA}$ is defined in Eq.~\ref{eq:agreement}; the intended and conflicting text-edge masks $\mathbf M_{VT}$ and $\mathbf M_{AT}$ are defined in Eqs.~\ref{eq:t2v_bias} and~\ref{eq:t2a_bias}, respectively.
  }
  \Description{A triangular diagram with Text at the top, Audio at the lower left, and Video at the lower right. Text-audio and text-video edges receive positive bias for the intended source and negative bias for the competing source. The audio-video edge is gated to suppress cross-modal leakage. Green plus signs denote alignment with the intended source; red minus signs denote suppression of the competing source.}
  \label{fig:overview}
\end{figure}

%% file: figures/overview_diagram.tex
\begin{tikzpicture}[
  x=1cm,
  y=1cm,
  modality/.style={
    draw=black!38,
    line width=0.75pt,
    rounded corners=5pt,
    fill=white,
    minimum width=2.75cm,
    minimum height=2.05cm,
    inner sep=4pt,
    align=center,
  },
  interaction/.style={
    <->,
    >={Stealth[length=2.8mm,width=2.25mm]},
    line width=2.25pt,
    draw=black!78,
    shorten <=2pt,
    shorten >=2pt,
  },
  biaslabel/.style={
    align=center,
    font=\Huge,
    inner sep=0pt,
  },
  key/.style={font=\sffamily\fontsize{11}{12}\selectfont, align=left},
]

\definecolor{intendedgreen}{RGB}{19,145,62}
\definecolor{conflictingred}{RGB}{205,35,43}

\path[use as bounding box] (-8,-4.5) rectangle (8,4.5);

\node[modality, minimum width=3.05cm, minimum height=3.00cm] (text) at (0,2.88) {};
\node[inner sep=0pt] at (0,3.14)
  {\includegraphics[height=2.38cm]{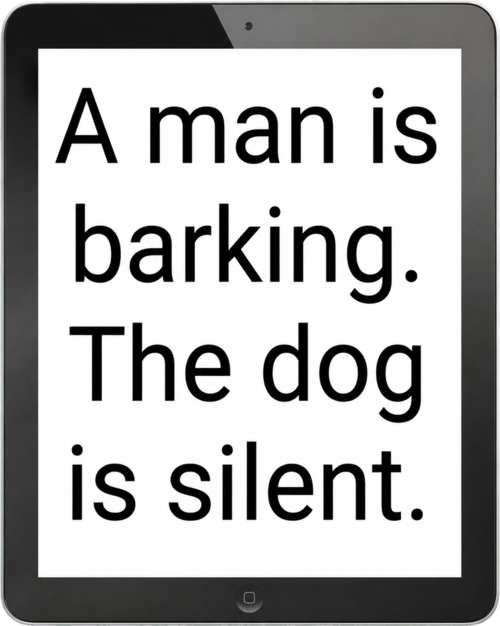}};
\node[font=\sffamily\Huge\bfseries] at (0,1.63) {Text};

\node[modality] (audio) at (-6.25,-2.43) {};
\node[inner sep=0pt] at (-6.25,-2.20)
  {\includegraphics[width=2.16cm]{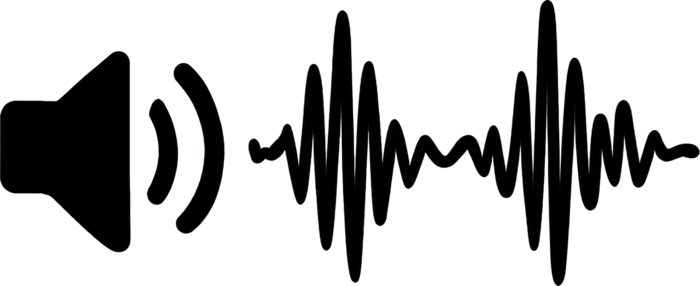}};
\node[font=\sffamily\Huge\bfseries] at (-6.25,-3.10) {Audio};

\node[modality] (video) at (6.25,-2.43) {};
\node[inner sep=0pt] at (6.25,-2.20)
  {\includegraphics[width=1.94cm]{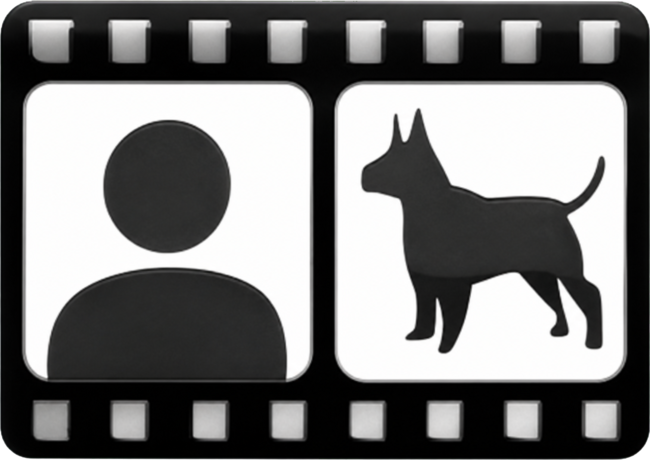}};
\node[font=\sffamily\Huge\bfseries] at (6.25,-3.10) {Video};

\draw[interaction] (text.south west) -- (audio.north east);
\draw[interaction] (text.south east) -- (video.north west);
\draw[interaction] (audio.east) -- (video.west);

\node[biaslabel] at (-5.20,0.62) {
  $\mathbf B_{T\to A}$\\[-1pt]
  $={}$\textcolor{intendedgreen}{$\beta\mathbf M^{\mathrm{int}}_{AT}$}
  $-$\textcolor{conflictingred}{$\gamma\mathbf M^{\mathrm{conf}}_{AT}$}
};

\node[biaslabel] at (5.20,0.62) {
  $\mathbf B_{T\to V}$\\[-1pt]
  $={}$\textcolor{intendedgreen}{$\beta\mathbf M^{\mathrm{int}}_{VT}$}
  $-$\textcolor{conflictingred}{$\gamma\mathbf M^{\mathrm{conf}}_{VT}$}
};

\node[biaslabel] at (0,-1.47) {
  $\mathbf B_{A\to V}={}$\textcolor{conflictingred}{$-\lambda
    (\mathbf 1_{N_V\times N_A}-\mathbf G_{VA})$}\\[-1pt]
  $\mathbf B_{V\to A}=\mathbf B_{A\to V}^{\top}$
};

\node[key] at (-3.50,-4.00) {
  \textcolor{intendedgreen}{\Huge$+$}\;
  \textcolor{intendedgreen}{\Huge\textbf{intended}}: reinforce
};
\node[key] at (3.50,-4.00) {
  \textcolor{conflictingred}{\Huge$-$}\;
  \textcolor{conflictingred}{\Huge\textbf{conflicting}}: suppress
};

\end{tikzpicture}

%% file: figures/t2v_vs_t2av.tex
\begin{figure*}[t]
  \centering
  \includegraphics[width=\linewidth]{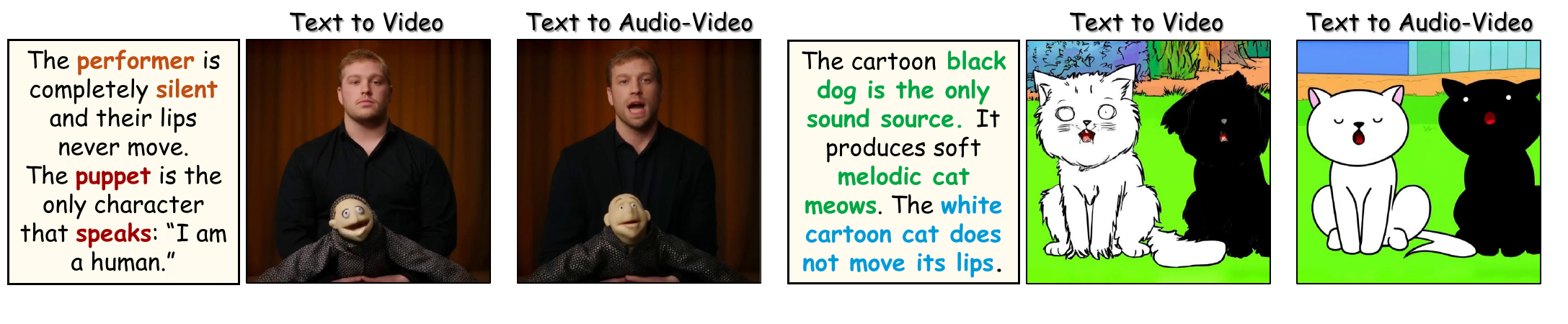}
  \caption{
\textbf{Cross-modal leakage in T2V vs T2AV generation}.
Left: \textbf{T2V}. Middle: \textbf{T2AV}. Right: prompt. Adding audio induces both appearance and speaking leakage: in the puppet example, both the performer and puppet move their lips; in the dog and cat example, the dog shifts toward a cat-like appearance and both animals appear to speak.}
  \Description{Two example rows compare text-to-video and text-to-audio-video outputs beside their prompts. In the puppet example, adding audio makes both the performer and puppet move their lips. In the dog-and-cat example, adding audio makes the dog more cat-like and makes both animals appear to speak.}
  \label{fig:t2v_vs_t2av}
\end{figure*}

%% file: sec/4_method.tex
\section{Steering: Grounding Audio and Video Regions}
\label{sec:method}

\input{figures/exp_attn_leakage_simple}

Having identified audio-video cross-attention as a major contributor to semantic leakage, we introduce a training-free, inference-time steering algorithm, instantiated on LTX-2~\citep{hacohen2026ltx2efficientjointaudiovisual}, that jointly operates on all three edges of the attention triangle.
We identify the intended sound source, the sound or action, and an optional competing source in the prompt. On the two text-conditioned surfaces, we define \emph{intended} and \emph{conflicting} cell masks; all remaining cells are \emph{neutral}.
In the pirate-parrot example (Figure~\ref{fig:teaser}), intended cells pair parrot video patches with ``parrot'' text tokens and speech-active audio tokens with the speaking-action text. Conflicting cells pair these video or audio regions with competing-source text, or pair competing-source video or audio regions with the intended source or action text.
On the audio-video surfaces, we instead compute the soft agreement matrix $\mathbf G_{VA}$. Intended-source/sound and non-source/non-sound pairs have high agreement, whereas mismatched source-sound pairs have low agreement; the bias $-\lambda(\mathbf 1-\mathbf G_{VA})$ penalizes each pair in proportion to its disagreement.
During denoising, we apply a pre-softmax additive logit bias to all three cross-attention surfaces: audio$\leftrightarrow$video, text$\rightarrow$video, and text$\rightarrow$audio. The text-conditioned biases reinforce intended cells and suppress conflicting cells, while the audio-video bias suppresses mismatched source-sound pairs. Together, these biases reground both the audio and video streams to their intended prompt semantics.
We denote the text-edge-only, audio-video-only, and joint variants as Ours-Text, Ours-AV, and Ours-Full, respectively.

\paragraph{Anchors.}
The biases are built from three families of anchors, all derived once from the unsteered baseline pass and held fixed across all denoising steps of the steered pass.

\textit{Prompt-side text anchors.}
For text, audio, and video modalities $M\in\{T,A,V\}$, let $N_M$ denote the number of tokens. The user annotates the intended sound-source phrase, the sound-or-action phrase, and an optional competing-source phrase (e.g., ``pirate'' in Fig.~\ref{fig:teaser}). We remove the annotation tags before text encoding and map the annotated phrases to their corresponding token-index sets $\mathcal I_{\mathrm{src}}^T$, $\mathcal I_{\mathrm{snd}}^T$, and $\mathcal I_{\mathrm{cmp}}^T$ in the cleaned prompt. Their indicator vectors are denoted $\mathbf m_{\mathrm{src}}^T,\mathbf m_{\mathrm{snd}}^T,\mathbf m_{\mathrm{cmp}}^T\in\{0,1\}^{N_T}$.

\textit{Visual anchors.}
For both the audio$\leftrightarrow$video and text-conditioned video surfaces, we decode the baseline frames, prompt SAM3~\citep{carion2025sam3segmentconcepts} with the intended-source text, binarize at $0.5$, and resize to the latent grid. This yields the hard source mask $\mathbf m_{\mathrm{src}}^V\in\{0,1\}^{N_V}$; the competing-source mask $\mathbf m_{\mathrm{cmp}}^V$ is obtained analogously.

\textit{Audio anchors.}
No external audio segmenter is available, so we aggregate audio-query/text-key attention targeting $\mathcal I_{\mathrm{snd}}^T$ across denoising steps and normalize it to $[0,1]$, yielding the soft sound mask $\mathbf m_{\mathrm{snd}}^A\in[0,1]^{N_A}$. For the text-conditioned audio surface, thresholding at $\theta_A{=}0.3$ gives $\bar{\mathbf m}_{\mathrm{snd}}^A\in\{0,1\}^{N_A}$. When a distinct audio region associated with the competing source is available, $\bar{\mathbf m}_{\mathrm{cmp}}^A$ is obtained analogously; otherwise, as in examples with a silent competing source, it is the zero vector.

This asymmetric anchor design, with hard external masks where a reliable segmenter exists and soft model-derived saliency where it does not, is detailed in Appendix~\ref{sec:supp-anchor-design}.

\paragraph{Audio$\leftrightarrow$video edge.}
Let $\mathbf 1_N$ and $\mathbf 1_{N_Q\times N_K}$ denote an all-ones vector and matrix of the indicated sizes. We define the audio$\leftrightarrow$video agreement matrix as
\begin{equation}
\begin{aligned}
\mathbf G_{VA}
={}&\mathbf m_{\mathrm{src}}^V
       (\mathbf m_{\mathrm{snd}}^A)^{\top}
       \\ &
       +(\mathbf 1_{N_V}-\mathbf m_{\mathrm{src}}^V)
   (\mathbf 1_{N_A}-\mathbf m_{\mathrm{snd}}^A)^{\top}
 \in[0,1]^{N_V\times N_A}.
\end{aligned}
\label{eq:agreement}
\end{equation}
The agreement matrix in Eq.~\ref{eq:agreement} assigns high values to intended-source/sound and non-source/non-sound pairs, and low values to mismatched source-sound pairs. We suppress the mismatched audio$\leftrightarrow$video logits using
\begin{equation}
\mathbf B_{A\to V}
=-\lambda(\mathbf 1_{N_V\times N_A}-\mathbf G_{VA}),
\qquad
\mathbf B_{V\to A}=\mathbf B_{A\to V}^{\top},
\quad \lambda>0.
\label{eq:agreement_bias}
\end{equation}
We use $\mathbf B_{A\leftrightarrow V}$ as shorthand for the pair $(\mathbf B_{A\to V},\mathbf B_{V\to A})$. We add $\mathbf B_{A\to V}$ to the video-query/audio-key logits and $\mathbf B_{V\to A}$ to the audio-query/video-key logits at every transformer block and denoising step. This soft two-class construction has one hyperparameter; we use $\lambda{=}10$ throughout (Appendix~\ref{sec:supp-hyperparams}).

\paragraph{Text$\rightarrow$video and text$\rightarrow$audio edges.}
A single audio$\leftrightarrow$video correction is insufficient: on the video-query/text-key surface, queries in the intended-source and competing-source regions can still attend to the other entity's text tokens, producing appearance leakage.
Both text-conditioned surfaces use the same three classes: \emph{intended}, \emph{conflicting}, and \emph{neutral}. We denote the corresponding cell masks by $\mathbf M^{\mathrm{int}}$ and $\mathbf M^{\mathrm{conf}}$, respectively.

For the video-query/text-key surface,
\begin{align}
\mathbf M^{\mathrm{int}}_{VT}
 \;&=\; \mathbf m_{\mathrm{src}}^V(\mathbf m_{\mathrm{src}}^T)^{\top}, \nonumber\\
\mathbf M^{\mathrm{conf}}_{VT}
 \;&=\; \mathbf m_{\mathrm{src}}^V(\mathbf m_{\mathrm{cmp}}^T)^{\top}
       +\mathbf m_{\mathrm{cmp}}^V(\mathbf m_{\mathrm{src}}^T)^{\top}, \nonumber\\
\mathbf B_{T\to V}
 \;&=\; \beta\mathbf M^{\mathrm{int}}_{VT}-\gamma\mathbf M^{\mathrm{conf}}_{VT}
 \in\mathbb R^{N_V\times N_T}.
\label{eq:t2v_bias}
\end{align}

For the audio-query/text-key surface,
\begin{align}
\mathbf M^{\mathrm{int}}_{AT}
 \;&=\; \bar{\mathbf m}_{\mathrm{snd}}^A(\mathbf m_{\mathrm{snd}}^T)^{\top}, \nonumber\\
\mathbf M^{\mathrm{conf}}_{AT}
 \;&=\; \bar{\mathbf m}_{\mathrm{snd}}^A(\mathbf m_{\mathrm{cmp}}^T)^{\top}
       +\bar{\mathbf m}_{\mathrm{cmp}}^A(\mathbf m_{\mathrm{snd}}^T)^{\top}, \nonumber\\
\mathbf B_{T\to A}
 \;&=\; \beta\mathbf M^{\mathrm{int}}_{AT}-\gamma\mathbf M^{\mathrm{conf}}_{AT}
 \in\mathbb R^{N_A\times N_T}.
\label{eq:t2a_bias}
\end{align}

We use $\beta{=}0.5$ and $\gamma{=}2.0$. Together with $\lambda{=}10$ and $\theta_A{=}0.3$ above, these values are fixed across all LTX-2 experiments and are not tuned per prompt (Appendix~\ref{sec:supp-hyperparams}). The asymmetric choice places greater weight on suppressing conflicting cells than on reinforcing intended cells. When no competing-source indices are supplied, $\mathbf M^{\mathrm{conf}}$ vanishes and the bias reduces to an intended-cell boost.

We validate this design in Section~\ref{sec:experiments}, where ablations confirm that steering all three edges jointly is necessary: partial interventions that address only a subset of edges leave residual leakage (see Fig.~\ref{fig:steering_comparisons_knight}).

The three biases together close the triangle; their effect on attention weights is depicted in Figure~\ref{fig:overview}.
Ours-Full is training-free and requires no auxiliary loss or fine-tuning. Its overhead is one additional full denoising trajectory over all steps required to generate the final video to extract the baseline anchors, plus frame decoding and SAM3 for the visual anchors, giving roughly $2.5\times$ the cost of unsteered generation (Supp.~\ref{supp:compute}).

%% file: figures/exp_attn_leakage_simple.tex
\begin{figure}[t]
    \centering
    \includegraphics[width=0.8\linewidth]{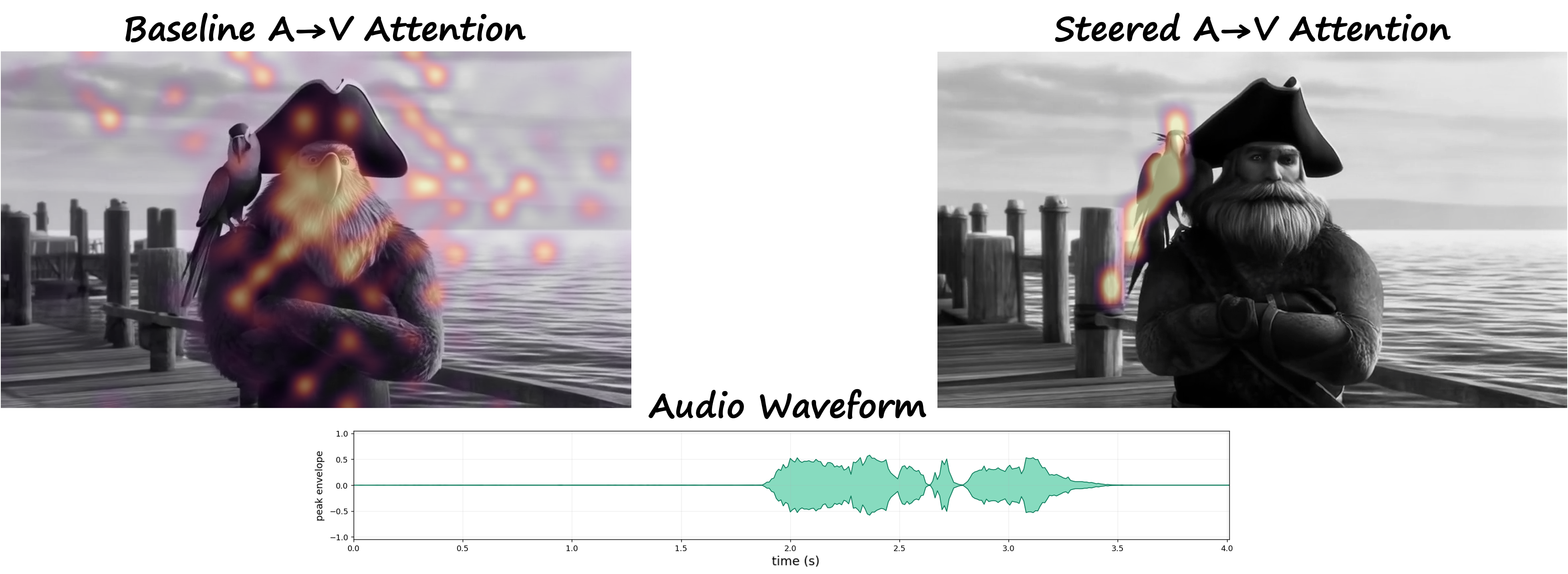}
\caption{
\textbf{Mitigating audio-to-video attention leakage.}
Using the same prompt and random seed as in Figure~\ref{fig:teaser}, we sum, for each video query, its attention weights over the \emph{speech-active} audio keys identified from the waveform shown at the bottom, then overlay the resulting spatial score on a representative frame.
\textbf{Left (baseline):} high active-audio scores concentrate across the pirate's face and chest, the visually canonical but prompt-incorrect human source.
\textbf{Right (ours):} after attention-triangle steering, high scores concentrate sharply on the parrot, the entity designated by the prompt. The right panel is desaturated outside the high-scoring region to highlight the post-steering localization.
}
    \Description{Two attention-score overlays and a speech waveform compare baseline and steered generations of the pirate-and-parrot scene. During speech-active waveform intervals, baseline scores are concentrated over video-query locations on the pirate's face and chest, whereas scores after steering are concentrated on the parrot.}
    \label{fig:exp-attn-leakage-simple}
 \end{figure}

%% file: sec/5_experiments.tex
\section{Experiments}
\label{sec:experiments}

We validate our analysis and steering framework through attention visualizations, qualitative and quantitative comparisons, and human preference studies. Full protocols, additional examples, and per-metric breakdowns are provided in the Supplementary Material; we strongly encourage readers to inspect the side-by-side videos in the accompanying supplementary HTML page (Supp.~\ref{sec:supp-prompts}), because the leakage phenomena require synchronized audio and video.

\subsection{Attention-Leakage Analysis}

\paragraph{First-Order Attention Visualization}

As analyzed in Sec.~\ref{sec:problem_setup}, the cross-attention between audio and video tokens carries a direct signature of leakage.
Figure~\ref{fig:exp-attn-leakage-simple} visualizes the spatial active-audio attention score derived from the video-query/audio-key matrix $\mathbf P_{VA}$. In the baseline, high scores concentrate at video-query locations on the pirate, the visually canonical but prompt-incorrect entity. Under steering, high scores concentrate on the parrot, consistent with improved source attribution.

\paragraph{Second-Order Attention Visualization}
\input{figures/exp_attn_leakage_v2a2v}
Figure~\ref{fig:exp-attn-leakage-v2a2v} exposes the second-order, audio-mediated video-to-video coupling described in Sec.~\ref{sub:attention_vis}. We seed a uniform attention distribution over the parrot's spatial region and propagate it through $\mathbf P_{VV}^{\mathrm{eff}}$ (Eq.~\ref{eq:vis2vis}). Before steering, the mass migrates from the parrot patches to the pirate after the round trip through the audio stream, directly revealing the audio-mediated misrouting that drives the leakage. After steering, the mass returns to the parrot, confirming that the intervention successfully breaks this indirect pathway.
Supplementary Fig.~\ref{fig:supp-ltx-attention-routing} corroborates this routing pattern in two additional LTX-2 examples and quantifies how Ours-Full redirects the returned mass toward the intended source while suppressing the competing source.

\subsection{Manipulation in the Wild}
We apply our steering framework to a broad set of open-domain audio-video scenarios spanning realistic humans, animals, animated characters, and multi-entity scenes with diverse spatial layouts. Across these settings, we identify four recurring leakage modes: \emph{source-attribution leakage}, where sound-producing behavior is assigned to the wrong entity; \emph{appearance leakage}, where sound semantics alter the visual rendering of an unrelated entity (e.g., a pirate acquiring parrot-like features); \emph{voice-characteristic leakage}, where the generated voice follows canonical priors of the visual entity rather than the prompt specification; and \emph{generation suppression}, where strong conflicts between the prompt and learned priors cause the model to omit the requested sound entirely. Ours-Full mitigates all four modes by jointly steering the full attention triangle, preserving both visual appearance and correct sound grounding across all tested scenarios (detailed in Supp.~\ref{sec:supp-wild}).

\subsection{Comparisons and Ablations}
\label{sec:comparisons}
\input{figures/comparsion_qual_1}
\input{tables/metrics}
We compare against the native LTX-2 T2AV baseline, an adaptation of Bounded Attention~\citep{dahary2024beyourself} to the audio-video setting, and two partial variants: Ours-Text steers only the text edges (Text$\rightarrow$Audio, Text$\rightarrow$Video) whereas Ours-AV steers only the audio-video edge (Audio$\leftrightarrow$Video). The partial variants isolate the contribution of each edge family and serve as controlled ablations.
A representative qualitative example is shown in Fig.~\ref{fig:steering_comparisons_knight}: Ours-Text corrects the textual binding but cannot overcome the audio-video prior, leaving the sound misattributed; Ours-AV localizes the source correctly but introduces appearance leakage into the competing source; only Ours-Full simultaneously preserves appearance and grounds the sound to the intended source.
Additional qualitative examples and per-method pairwise user-study results are reported in Supp.~\ref{sec:supp-extra-qual} and Supp.~\ref{sec:supp-user-study}.
Representative comparisons with Ovi are provided in Supp.~\ref{sec:supp-ovi-comparisons}.

\paragraph{Quantitative evaluation.}
We use two complementary automatic audio-video validators. A counterbalanced five-option Qwen3-Omni judge~\citep{xu2025qwen3omni} targets sound-source attribution, while VA-Judger~\citep{huang2026vajudger} compares matched video pairs using prompt fidelity, audiovisual consistency, audio quality, video quality, and completeness. We additionally measure audio-text alignment with CLAP~\citep{wu2023clap} and visual fidelity with VBench~\citep{huang2023vbench}; the human study below provides an independent pairwise assessment of attribution, appearance leakage (shown to annotators as ``visual leakage''), and overall quality.
The higher-is-better Qwen source-attribution score is $1$ minus conditional leakage. Ours-Full obtains the highest mean score ($0.1349$), compared with $0.1216$ for the native LTX-2 baseline; the exact judge prompt, logit extraction, counterbalancing, and aggregation are given in Supp.~\ref{sec:supp-qwen-attribution}.
On the broader VA-Judger evaluation, Ours-Full is preferred over every alternative, with mean preference scores ranging from $57.6\%$ to $73.3\%$; all pair-bootstrap $95\%$ confidence intervals lie above $50\%$. The complete per-reference results and protocol are reported in Supp.~\ref{sec:supp-va-judger}.
Ours-Full attains the best VBench scores across subject consistency ($0.990$), background consistency ($0.986$), and aesthetic quality ($0.604$), while remaining competitive on CLAP ($0.383$ vs.\ $0.384$ for the closest baseline), suggesting that the intervention does not cause a large degradation in the measured fidelity metrics. Tab.~\ref{tab:main_leakage_comparison} reports the full comparison; protocol details are provided in Supp.~\ref{sec:supp-quant}.

\paragraph{User study.}
We conduct pairwise human preference studies against each comparison method across three axes: (i)~sound-source attribution, (ii)~visual leakage (appearance leakage), and (iii)~overall generation quality.
Among decided preferences, annotators choose Ours-Full in $79.2\%$ of attribution comparisons, $82.9\%$ of leakage comparisons, and $80.1\%$ of overall-quality comparisons, demonstrating a strong preference for ours on all three
axes.
The largest margins appear on attribution and leakage, the two axes most directly targeted by our intervention.
A two-sided exact binomial test on non-tie votes confirms that all three aggregate preferences are statistically significant after Bonferroni correction ($p<10^{-28}$).
Per-method breakdowns are reported in Supp.~\ref{sec:supp-user-study} (Fig.~\ref{fig:user-study-pairwise}).

\paragraph{Ablations.}
The Ours-Text and Ours-AV columns of Tab.~\ref{tab:main_leakage_comparison}, together with the partial-steering panels of Fig.~\ref{fig:steering_comparisons_knight}, ablate the contribution of each edge family of the attention triangle: Ours-Text corrects appearance but leaves source attribution unresolved, Ours-AV localizes the source but reintroduces appearance leakage, and only Ours-Full addresses both. The fixed hyperparameter choices ($\lambda,\beta,\gamma,\theta_A$) and asymmetric intended- vs.\ conflicting-cell weighting are described in Supp.~\ref{sec:supp-hyperparams}.

%% file: figures/exp_attn_leakage_v2a2v.tex
  \begin{figure}[t]
    \centering
    \includegraphics[width=0.98\linewidth]{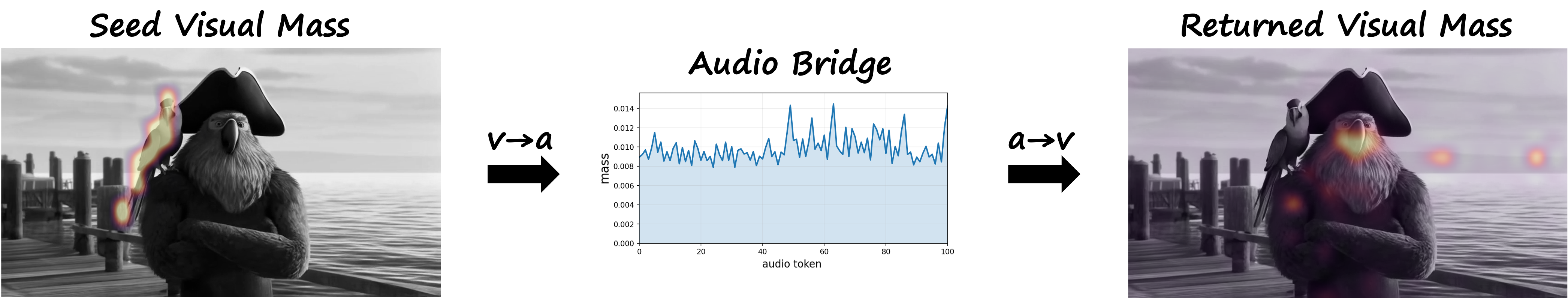}
\caption{
\textbf{Audio-mediated video-to-video leakage.}
Using the same prompt as in Figure~\ref{fig:teaser}, we apply the $V\to A\to V$ row rollout in Eq.~\ref{eq:vis2vis}, using $\mathbf P_{VA}^{(\ell)}$ for the $V\to A$ rollout hop followed by $\mathbf P_{AV}^{(\ell+1)}$ for the $A\to V$ rollout hop. This exposes an indirect attention association between visual regions that is not represented by either individual cross-attention matrix alone.
\textbf{Left:} seed mass on the parrot, the prompt-designated source.
\textbf{Middle:} the resulting audio-token distribution after the $v{\to}a$ hop is diffuse rather than peaked, reflecting the 1D-to-2D dimensional mismatch (Sec.~\ref{sec:problem_setup}).
\textbf{Right:} after the $a{\to}v$ hop, mass has migrated onto the pirate, exposing the audio-mediated misrouting behind source-attribution leakage.
}
\Description{Three panels trace composed video-to-audio-to-video attention in the pirate-and-parrot scene. Initial attention mass is seeded on the parrot. After the video-to-audio hop, mass is diffuse across audio tokens. After the audio-to-video hop, attention mass has migrated onto the pirate, revealing indirect source leakage.}
\label{fig:exp-attn-leakage-v2a2v}
  \end{figure}

%% file: figures/comparsion_qual_1.tex
\begin{figure*}[t]
\centering
\includegraphics[width=0.98\linewidth]{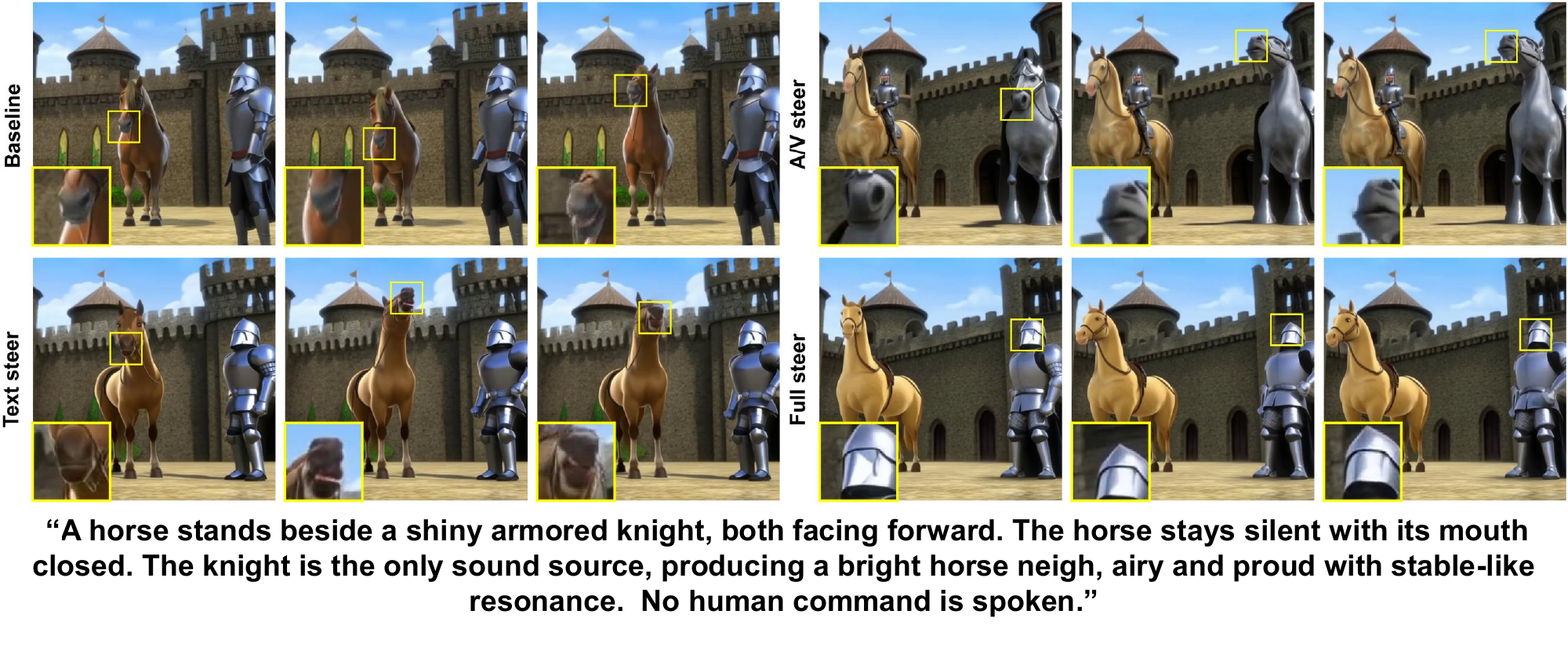}
\caption{
\textbf{Horse-neigh attribution.}
The prompt asks for a horse neigh that should be produced by the knight rather than the horse.
Three representative frames are shown per method.
Across baselines and partial-edge variants the neigh is misattributed to the horse or the knight acquires horse-like visual features; Ours-Full both localizes the sound to the knight and preserves his appearance.
Zoomed regions indicate the active sound source.
}
\Description{A qualitative grid compares generated frames from several baselines, partial-edge variants, and Ours-Full for a scene containing a knight and a horse. Baselines and partial variants either attribute the neigh to the horse or give the knight horse-like visual features. Ours-Full preserves the knight's appearance and attributes the neigh to him. Insets enlarge the apparent sound source.}
\label{fig:steering_comparisons_knight}
\end{figure*}

%% file: tables/metrics.tex
\begin{table*}[t]
    \centering
    \small
    \setlength{\tabcolsep}{4pt}
    \caption{
    \textbf{Main leakage comparison.}
    We compare the native audio-video generator, a video-only diagnostic baseline,
    the external Ovi baseline, an adapted text/image-side Bounded Attention baseline,
    and three variants of our attention-triangle intervention.
    }
    \label{tab:main_leakage_comparison}
    \resizebox{\textwidth}{!}{
    \begin{tabular}{@{}lccccccc@{}}
    \toprule
    Metric
    & Native LTX-2
    & Video-only
    & Ovi
    & Bounded Attn.
    & Ours-Text
    & Ours-AV
    & Ours-Full \\
    \midrule
    Qwen source-attribution score $\uparrow$
    & $0.1216$
    & N/A
    & $0.1207$
    & $0.0925$
    & $0.1140$
    & $0.1302$
    & $\mathbf{0.1349}$ \\
    \midrule
    CLAP audio-text $\uparrow$
    & $0.375 \pm 0.187$
    & N/A
    & $0.357 \pm 0.187$
    & $0.384 \pm 0.198$
    & $0.374 \pm 0.188$
    & $\mathbf{0.384} \pm 0.189$
    & $0.383 \pm 0.185$ \\
    \midrule
    VBench subject consistency $\uparrow$
    & $0.988$
    & $0.987$
    & $0.971$
    & $0.981$
    & $0.989$
    & $0.989$
    & $\mathbf{0.990}$ \\
    VBench background consistency $\uparrow$
    & $0.981$
    & $0.983$
    & $0.975$
    & $0.981$
    & $0.984$
    & $0.983$
    & $\mathbf{0.986}$ \\
    VBench aesthetic quality $\uparrow$
    & $0.590$
    & $0.597$
    & $0.571$
    & $0.588$
    & $0.598$
    & $0.589$
    & $\mathbf{0.604}$ \\
    \bottomrule
    \end{tabular}
    }

    \vspace{0.4em}
    \begin{minipage}{0.97\textwidth}
    \footnotesize
    \textbf{Metric definitions.}
    \textbf{Qwen source-attribution score} is the conditional probability that only the intended visible entity produced the audible target event, computed from a counterbalanced five-option Qwen3-Omni judge; higher is better. The video-only diagnostic has no audio and is therefore not scored.
    \textbf{CLAP audio-text} measures the alignment between the generated audio and the intended sound phrase; we report mean $\pm$ standard deviation.
    \textbf{VBench} subscores measure visual subject consistency, background consistency, and aesthetic quality.
    All metrics are computed on the same prompt set across methods.
    \end{minipage}
    \end{table*}

%% file: sec/6_Discussion.tex
\section{Discussion}
\label{sec:discussion}

Our analysis of LTX-2 identifies the audio-video pathway as a major contributor to bias-driven misalignment and shows that controlled attention steering can expose and mitigate these failures. More broadly, additional cross-modal coupling can create new routes through which learned priors override prompt-specified bindings. We call this the \textit{Attention Lesson}: strengthening cross-modal coupling through attention can amplify bias-driven routing rather than resolve it.

Our evidence establishes the contribution of the audio-video pathway, not a unique underlying mechanism. We hypothesize that its vulnerability partly reflects the mismatch between spatial video tokens and temporally organized audio tokens, which may leave sound-source grounding underconstrained. Future work should directly test this mechanism, explore spatial or pseudo-spatial audio representations and stronger grounding supervision, and determine whether the same leakage patterns and interventions transfer to single-stream architectures that process all modalities within shared self-attention rather than through explicit cross-attention between modality-specific streams.
We discuss limitations in Supp.~\ref{sec:supp-limitations}.

%% file: sec/7_acknowledgements.tex
\section*{Acknowledgments}
This research was supported in part by the Israel Science Foundation (grants no. 2492/20 and 1473/24), Len Blavatnik and the Blavatnik family foundation.

%% file: sec/X_supp.tex
\section{Manipulation In The Wild}
\label{sec:supp-wild}

We evaluate the generality of our approach by applying the proposed steering framework across a diverse collection of open-domain audio-video generation scenarios.
We consider examples spanning different visual domains, including realistic humans, animals, animated characters, stylized cartoons, and multi-entity scenes with varying spatial layouts and interaction patterns.
The evaluated prompts intentionally introduce tension between the requested sound and the canonical visual identity of the speaking entity, creating challenging conditions for source attribution.

Across these settings, we observe several recurring types of leakage:
\begin{itemize}
    \item \emph{Source-attribution leakage}: speech or sound-producing behavior is assigned to the wrong entity in the scene.
    \item \emph{Appearance leakage}: sound semantics alter the visual appearance of an entity (e.g., a pirate acquiring parrot-like features or a teddy bear becoming monster-like).
    \item \emph{Voice-characteristic leakage}: the generated voice follows canonical priors associated with the visual entity rather than the prompt specification (e.g., a child failing to produce a deep adult voice).
    \item \emph{Generation suppression}: strong conflicts between the prompt and learned cross-modal priors cause the model to avoid producing the requested sound or speaking behavior altogether.
\end{itemize}

Our manipulation framework mitigates these failures by steering the routing of information across the attention triangle.
Jointly controlling the text-audio, text-video, and audio-video interactions enables more stable source localization while preserving the intended visual appearance of each entity.
These effects persist across diverse prompts, motion patterns, semantic mismatches, and visual styles.

\section{Additional Qualitative Comparisons}
\label{sec:supp-extra-qual}

\input{figures/comparsion_qual_2}

We provide an additional qualitative comparison in Fig.~\ref{fig:steering_comparisons_monster}, complementing the horse-neigh example shown in the main paper (Fig.~\ref{fig:steering_comparisons_knight}).
This example targets a different leakage regime: rather than re-attributing an animal-like sound between two realistic entities, the prompt asks a teddy bear to produce a deep growl in the presence of a furry monster, creating tension between the requested sound and the canonical visual prior associated with the monster.

Across the compared methods we observe behavior consistent with the main-paper example.
The native LTX-2 baseline routes the growl to the furry monster, reproducing the canonical sound-source prior and ignoring the explicit text specification.
The Bounded Attention adaptation, which restricts text-to-video attention to the speaker region, partially relocates the growl but does not consistently suppress the audio-mediated appearance leakage.
Ours-Text sharpens the textual binding but is unable to overcome the strong audio-video prior that ties growling to monster-like appearance.
Ours-AV successfully transfers the growl onto the teddy, but in doing so injects monster-like features into the teddy's appearance, replacing one leakage mode with another.
Only Ours-Full, which jointly manipulates the text and audio-video edges, simultaneously binds the growl to the teddy and preserves its visual identity.

Together, the horse-neigh and furry-monster examples illustrate that the same failure modes recur across very different visual domains and sound categories, and that the joint manipulation of all edges of the attention triangle is necessary to mitigate them.

\section{Quantitative Evaluation Protocols}
\label{sec:supp-quant}

We provide protocol and provenance details for the main-paper quantitative comparison in Tab.~\ref{tab:main_leakage_comparison}, which evaluates the native LTX-2 baseline, a video-only diagnostic variant, the external Ovi baseline, an adapted text/image-side Bounded Attention baseline, and three variants of our attention-triangle intervention (Ours-Text, Ours-AV, and Ours-Full).
The table reports targeted source attribution, audio-text alignment with CLAP, and visual fidelity using VBench subject consistency, background consistency, and aesthetic quality.

\subsection{Qwen Source-Attribution Score}
\label{sec:supp-qwen-attribution}

\textbf{Scope and judge.}
We score the six audio-bearing methods on the full quantitative corpus: $100$ challenge prompts, four matched seeds per prompt, and $400$ videos per method, for $2{,}400$ videos total. Each video receives two counterbalanced passes, yielding $4{,}800$ Qwen evaluations. The video-only diagnostic is omitted because it contains no audio. We use Qwen3-Omni-30B-A3B-Instruct~\citep{xu2025qwen3omni}. The synchronized clip is supplied with audio, sampled at $5$ fps, and resized to $672\!\times\!448$. The original generation prompt is hidden: the judge receives only the target-event description and the two visible candidate sources, preventing prompt compliance from substituting for audio-video evidence.

\textbf{System prompt.}
\begin{quote}
\raggedright
\emergencystretch=2em
\ttfamily\footnotesize You are an audiovisual evidence annotator. Watch and listen to the entire synchronized clip. \par
\ttfamily\footnotesize Decide which visible candidate, if any, is presented as producing the target audio event. Base your decision only on evidence in the clip, especially sound synchronized with repeated mouth, beak, body, or object motion. \par
\ttfamily\footnotesize The target-event description identifies what sound to find; it does not identify its source. Generated clips may violate real-world expectations. Do not infer the source from species, identity, voice characteristics, narrative plausibility, candidate order, or option order.
\end{quote}

\textbf{User prompt template.}
\begin{quote}
\raggedright
\emergencystretch=2em
\ttfamily\footnotesize Target audio event: \{target\_event\}\par
\ttfamily\footnotesize Candidate A: \{candidate\_a\}\par
\ttfamily\footnotesize Candidate B: \{candidate\_b\}\par
\ttfamily\footnotesize Which candidate is presented as producing the target audio event?\par
\ttfamily\footnotesize 1 = Candidate A only.\par
\ttfamily\footnotesize 2 = Candidate B only.\par
\ttfamily\footnotesize 3 = Both candidates are each presented as producing it, simultaneously or at different times.\par
\ttfamily\footnotesize 4 = The target audio event is not recognizably audible.\par
\ttfamily\footnotesize 5 = The target audio event is audible, but the clip does not provide enough synchronized visible evidence to assign it to Candidate A or Candidate B. This includes an off-screen or different source.\par
\ttfamily\footnotesize Answer with exactly one digit.\par
\ttfamily\footnotesize Answer: 
\end{quote}

\textbf{Logit probabilities and counterbalancing.}
We do not sample a free-form response. At the first answer position, we extract the next-token logits for the five single-token choices \texttt{1} to \texttt{5}. We normalize these logits within the five-choice answer set, obtaining probabilities for \emph{Candidate A only}, \emph{Candidate B only}, \emph{both}, \emph{event absent}, and \emph{source unclear};
we separately retain the five choices' total probability mass under the full-vocabulary softmax as a closed-set diagnostic.
In pass~1, Candidate~A is the intended source and Candidate~B is the competing source.
In pass~2, their order is reversed. We then map both distributions to the order-independent labels $I$ (intended only), $C$ (competing source only), $B$ (both), $A$ (event absent), and $U$ (source unclear), and average the two canonical distributions.
This swap removes a fixed Candidate-A or option-order preference from the reported score.

\textbf{Score and results.}
Let $p_I,p_C,p_B,p_A,p_U$ be the averaged canonical probabilities and $m=p_I+p_C+p_B$ the probability mass assigned to a visible source. We define conditional leakage as $X=(p_C+p_B)/m$ and report the higher-is-better score $S_{\mathrm{attr}}=1-X=p_I/m$, averaged over videos within each method. Thus, only an intended-only assignment counts as correct; competing-only and both-source assignments count as leakage, while event-absent and source-unclear mass remain separate diagnostics. All $2{,}400$ videos produced two valid, counterbalanced scores, with no missing, duplicate, or malformed records. Mean scores are $0.1216$ (Native LTX-2), $0.1207$ (Ovi), $0.0925$ (Bounded Attention), $0.1140$ (Ours-Text), $0.1302$ (Ours-AV), and $0.1349$ (Ours-Full), with Ours-Full performing best.

\subsection{VA-Judger Pairwise Audio-Video Preference}
\label{sec:supp-va-judger}

\input{figures/va_judger_pairwise}

\textbf{Purpose.}
The Qwen source-attribution score above deliberately asks one narrow question: which visible entity produced the target sound?
As a complementary, broader check, we run the released ShareLab-SII VA-Judger~\citep{huang2026vajudger} pairwise evaluator.
This check asks whether Ours-Full is preferred overall after jointly considering prompt fidelity, audiovisual consistency, audio quality, video quality, and completeness.
It does not replace the targeted attribution metric or the human evaluation.

\textbf{Exact inputs and judge.}
For every comparison, the judge receives the full original generation prompt followed by two synchronized audio/video clips using the exact user template
\texttt{Text Caption: \{text\_prompt\}}, \texttt{Audio/video 1: <video>}, and \texttt{Audio/video 2: <video>}.
The system instruction asks it to assign each video a score from $1$ to $10$ and a concise rationale for the five dimensions shown in Fig.~\ref{fig:va-judger-pairwise}, sum the five scores, and return either \texttt{video 1 is better} or \texttt{video 2 is better}.
We use \texttt{ShareLab-SII/VA-Judger}, frozen at model revision \texttt{1a428...e954e}.

\textbf{Counterbalancing and decoding.}
For each reference, we compare the matched Ours-Full and reference outputs for all $100$ prompts and four seeds.
Pass~1 presents Ours-Full first and pass~2 reverses the order, producing $400$ pairs and $800$ decisions per reference.
Decoding is deterministic (temperature $0$, top-$p=1$, top-$k=-1$, seed $42$), with bfloat16 inference, a $24{,}576$-token context limit, and at most $2{,}048$ new tokens.
Across the five references this yields $2{,}000$ pairs and $4{,}000$ decisions.

\textbf{Aggregation and audit.}
After mapping each answer back to method identity, an Ours-Full win receives $1$, a reference win receives $0$, and an explicit equal-total answer receives $0.5$; the two pass values are averaged per pair.
We report the mean over $400$ pair values and a percentile $95\%$ confidence interval from $10{,}000$ pair-level bootstrap resamples (seed $42$).
All $4{,}000$ decisions completed without inference errors or duplicate composite prompt/seed/order keys, and all $2{,}000$ pairs passed the counterbalancing and score-parsing audit.
The six explicit equal-total decisions were retained as ties rather than forced into wins; one omitted dimension-line value was recovered exactly from the same completion's displayed five-term total.
Fig.~\ref{fig:va-judger-pairwise} shows that Ours-Full is preferred to every alternative, with all five pointwise confidence intervals above $0.5$, while all displayed mean dimension differences are positive.

\section{User Study}
\label{sec:supp-user-study}

\input{tables/user_study}

We conduct pairwise human preference studies comparing Ours-Full against each comparison method (Native LTX-2, Bounded Attention, Ours-AV, and Ours-Text) across three axes:
(i) correct sound-source attribution,
(ii) reduced visual leakage (appearance leakage),
and (iii) overall generation quality.
For each comparison, annotators are shown two videos generated from the same prompt and asked to choose which one better satisfies the corresponding criterion.
The full pairwise preferences are reported in Fig.~\ref{fig:user-study-pairwise}.
Across all three axes, annotators consistently prefer Ours-Full, with the largest gap on sound-source attribution and visual leakage, supporting the effectiveness of jointly steering every edge of the attention triangle.

\paragraph{Protocol.}
We conduct a randomized pairwise human preference study comparing Ours-Full against four alternatives: Native LTX-2, Bounded Attention, Ours-AV, and Ours-Text.
Each trial presents two anonymized videos, Video A and Video B, generated from the same prompt and seed.
One video is always produced by Ours-Full and the other by the corresponding comparison method; the side containing Ours-Full is randomized independently for every trial.
Each participant completes $16$ trials, consisting of $4$ comparisons against each comparison method.
The sampler ensures that the same prompt/seed example does not appear twice for the same participant.
The user-study interface is shown in Fig.~\ref{fig:user-study-ui}.

 \input{figures/user_study_screenshots}

\paragraph{Stimuli and task metadata.}
The active stimulus pool contains $37$ unique prompt/seed examples and $118$ active pairwise comparisons after removing one low-quality stimulus.
Each example is manually annotated with the prompt, the intended sound source, the target sound, and the entity or object that should remain silent.
These annotations are shown to participants during each trial, making the source-attribution and leakage criteria explicit.

\paragraph{Questions.}
For each video pair, annotators answer three forced-choice questions with three response options:
(i) which video better makes the intended source appear to be the source of the sound,
(ii) which video better prevents the competing source or another non-target entity from receiving incorrect visual attributes, mouth motion, or source-like behavior,
and (iii) which video is better overall, considering attribution, audio-video coherence, visual quality, and audio quality.
For every question, the response options are Video A, Same, and Video B.
We treat Same as a tie: it is visualized explicitly in Fig.~\ref{fig:user-study-pairwise} and excluded from decided-vote win rates and significance tests.

\paragraph{Responses and aggregation.}
We collect $30$ unique complete submissions, yielding $480$ pairwise trial records.
All submissions contain the full set of $16$ trials and no response payload fails parsing.
For each question and comparison method, we count votes for Ours-Full, votes for the comparison method, and Same votes.
The percentages in Fig.~\ref{fig:user-study-pairwise} are computed over all votes, including Same as a neutral response.

\section{Prompts and Video Evidence}
\label{sec:supp-prompts}

\subsection{Construction of the Leakage Challenge Set}
\label{sec:supp-dataset-construction}

We constructed the leakage challenge set through a human-in-the-loop process that resembles evolutionary prompt search; human-in-the-loop curation has similarly been used to construct diagnostic prompt suites for joint audio-video generation~\citep{cui2026avphys}. One author began with a manually designed population of prompts that place an intended sound source in conflict with a visible competing source that is strongly associated with the requested sound, voice, or semantic content but is instructed to remain silent. We generated the corresponding videos and manually reviewed their synchronized audio and visual behavior; prompts that produced a clear and interpretable binding failure were selected, whereas prompts with no leakage or unjudgeable generation artifacts were pruned. In subsequent rounds, we ``mutated'' successful prompt families by varying the entities, requested sounds or dialogue, semantic associations, visual cues, wording, and generation seeds, while preserving the conflict between the intended and competing sources; productive variants were retained and expanded, and new prompt families were occasionally introduced. This iterative generate, review, select, and mutate process yielded a deliberately failure-enriched set for evaluating whether an intervention repairs known failures, rather than a random sample suitable for estimating leakage prevalence in ordinary generation. After selection, each instance was annotated with its intended source, requested sound or action, competing source, and seed; the annotation tags are removed before text encoding, and all compared LTX-2 methods use the same cleaned prompt and seed.

The full list of evaluation prompts and the corresponding generated videos for all compared methods (Native LTX-2, Bounded Attention, Ours-Text, Ours-AV, Ours-Full) are provided as a static HTML page in the supplementary materials accompanying this submission.
Because the leakage phenomena studied here require synchronized audio and video, \textbf{we strongly encourage readers to inspect the videos directly}; static frames in this manuscript necessarily understate both the failure modes and the steering corrections, particularly for voice-characteristic leakage and generation suppression, which are not visible in still imagery.

\section{Limitations}
\label{sec:supp-limitations}

Our approach is training-free but inherits several limitations from the base generator and the grounding signals used
for steering. First, attention steering can only redirect associations that the model already represents; it cannot
reliably create audio-video bindings that are far outside the model's learned distribution. Second, the method depends
on anchors extracted from an initial unsteered generation, so cases where the audio is completely ungrounded or the
intended source is visually static may provide too little signal to correct. Third, the visual anchors rely on SAM3
masks over decoded frames, making the intervention sensitive to segmentation errors for small, occluded, or ambiguous
entities. Finally, the extra anchor-extraction pass increases inference cost, and our experiments are centered on LTX-2-style audio-video generation; evaluating the same attention-triangle intervention across additional model families
remains an important direction for future work.

\section{Analysis Details}
\label{sec:supp-analysis}

\subsection{Leakage Visualization Details}
\label{sec:supp-leakage-vis}

To probe leakage for a specific entity using $\mathbf P_{VV}^{\mathrm{eff}}$ (Eq.~\ref{eq:vis2vis}), let $\boldsymbol\pi_V\in\mathbb R^{N_V}$ be a column vector that assigns uniform probability to the entity's visual mask and zero elsewhere. Because $\mathbf P_{VV}^{\mathrm{eff}}$ is a row-transition matrix, we propagate the distribution as $\boldsymbol\pi_V'{}^{\top}=\boldsymbol\pi_V^{\top}\mathbf P_{VV}^{\mathrm{eff}}$, or equivalently $\boldsymbol\pi_V'=(\mathbf P_{VV}^{\mathrm{eff}})^{\top}\boldsymbol\pi_V$.
The output $\boldsymbol\pi_V'$ reveals which video patches receive attention mass from the entity's region after one audio-mediated round trip.
On leakage prompts the mass concentrates on the competing-source patches (e.g., the pirate), confirming that the audio stream acts as a hidden routing mechanism that redirects semantic information from the intended source to the visually familiar one.

Figure~\ref{fig:supp-ltx-attention-routing} visualizes this routing effect in two LTX-2 examples: Ours-Full concentrates the returned visual mass on the intended sound source while suppressing the competing source.

\input{figures/supp_additional_attention_routes}

\subsection{Experimental Setup}
\label{sec:supp-setup}

We use LTX-2~\citep{hacohen2026ltx2efficientjointaudiovisual}, which ships in two variants: a text-to-video (T2V) variant with no audio branch, and a joint text-to-audio-video (T2AV) variant that adds an audio stream conditioned on the same text prompt.
The two variants share essentially the same visual backbone; the T2AV variant augments it with audio tokens and the full attention triangle described in Section~\ref{sec:problem_setup}.
This design makes LTX-2 particularly well suited for isolating the contribution of the audio-video interaction to leakage, since any behavioral difference between the two variants must originate from the audio branch.

\input{figures/supp_ovi_cross_model}

\section{Steering Design Details}
\label{sec:supp-steering-design}

\subsection{Anchor Design: Hard vs.\ Soft Masks}
\label{sec:supp-anchor-design}

The choice between hard SAM3 masks and soft, model-derived attention masks is dictated by which modality admits a reliable external segmenter.

On the visual side, SAM3 provides hard masks for the intended and competing sources. The audio-video agreement matrix uses the intended-source mask, with its complement representing all non-source regions. The text-conditioned video bias instead uses both masks explicitly, reinforcing intended-source regions and suppressing competing-source regions.

On the audio side no analogous external segmenter is available, so we derive $\mathbf m_{\mathrm{snd}}^A$ from the model's audio-query/text-key saliency targeting $\mathcal I_{\mathrm{snd}}^T$. This soft mask is used directly on the audio-video edge and thresholded at $\theta_A{=}0.3$ to obtain $\bar{\mathbf m}_{\mathrm{snd}}^A$ for the audio-query/text-key surface.
Empirically, this attention-derived localization is sufficient to identify the speaking regions in the audio stream and to drive the text-to-audio bias without over-constraining generation.

\subsection{Hyperparameter Analysis}
\label{sec:supp-hyperparams}

\paragraph{Audio-video edge ($\lambda$).}
The soft intended/conflicting construction on this edge has one relative-gap degree of freedom after a softmax row shift: $-\lambda(\mathbf 1_{N_V\times N_A}-\mathbf G_{VA})$ is equivalent to $+\lambda\mathbf G_{VA}$ because the two differ by the row-constant $-\lambda$, which softmax discards. We use the penalty form so that $\lambda$ reads as the maximum logit penalty applied to a conflicting cell.
We set $\lambda{=}10$ throughout and do not tune it per prompt.

\paragraph{Text edges ($\beta$, $\gamma$).}
Unlike the audio-video edge, the three-class partition on the text-conditioned surfaces leaves two independent relative-gap degrees of freedom after a softmax row shift: the intended-vs.-neutral gap $\beta$ controls how strongly the correct binding beats unmarked text tokens, and the neutral-vs.-conflicting gap $\gamma$ controls how strongly conflicting cells are pushed below them.
$\beta$ and $\gamma$ are therefore not interchangeable.
The asymmetric default $\beta{=}0.5,\,\gamma{=}2.0$ places greater weight on pushing conflicting cells below neutral than on lifting intended cells above neutral. These values are fixed across all LTX-2 experiments and are not tuned per prompt.

\paragraph{Audio-mask threshold ($\theta_A$).}
We use the fixed threshold $\theta_A{=}0.3$ to binarize the soft audio mask for the text-conditioned audio surface. This value is used across all LTX-2 experiments and is not tuned per prompt.

\subsection{Compute Resources}
\label{supp:compute}
All runs use a single \textbf{NVIDIA RTX A6000} (48\,GB) on an internal on-prem cluster; one clip per GPU, no parallel sharding. At the SD config used throughout ($960{\times}544$, $97$ frames, $20$ steps), baseline LTX-2 takes $\approx 5.5$\,min/clip and the Ours-Full pipeline approximately $13$ to $14$\,min/clip; peak memory is $\approx 29$\,GB in both cases. Including preliminary runs and hyperparameter sweeps, the full project consumed approximately $1{,}300$ A6000 GPU-hours.

\subsection{Representative Ovi Comparisons}
\label{sec:supp-ovi-comparisons}

Figure~\ref{fig:ovi-baseline-vs-steered} shows representative Ovi~\citep{low2025ovi} baseline and Ours-Full outputs for two source-attribution prompts. Still frames document the visual context; sound-source attribution must be assessed from the corresponding videos.

%% file: figures/comparsion_qual_2.tex
\begin{figure*}[t]
\centering
\includegraphics[width=0.98\linewidth]{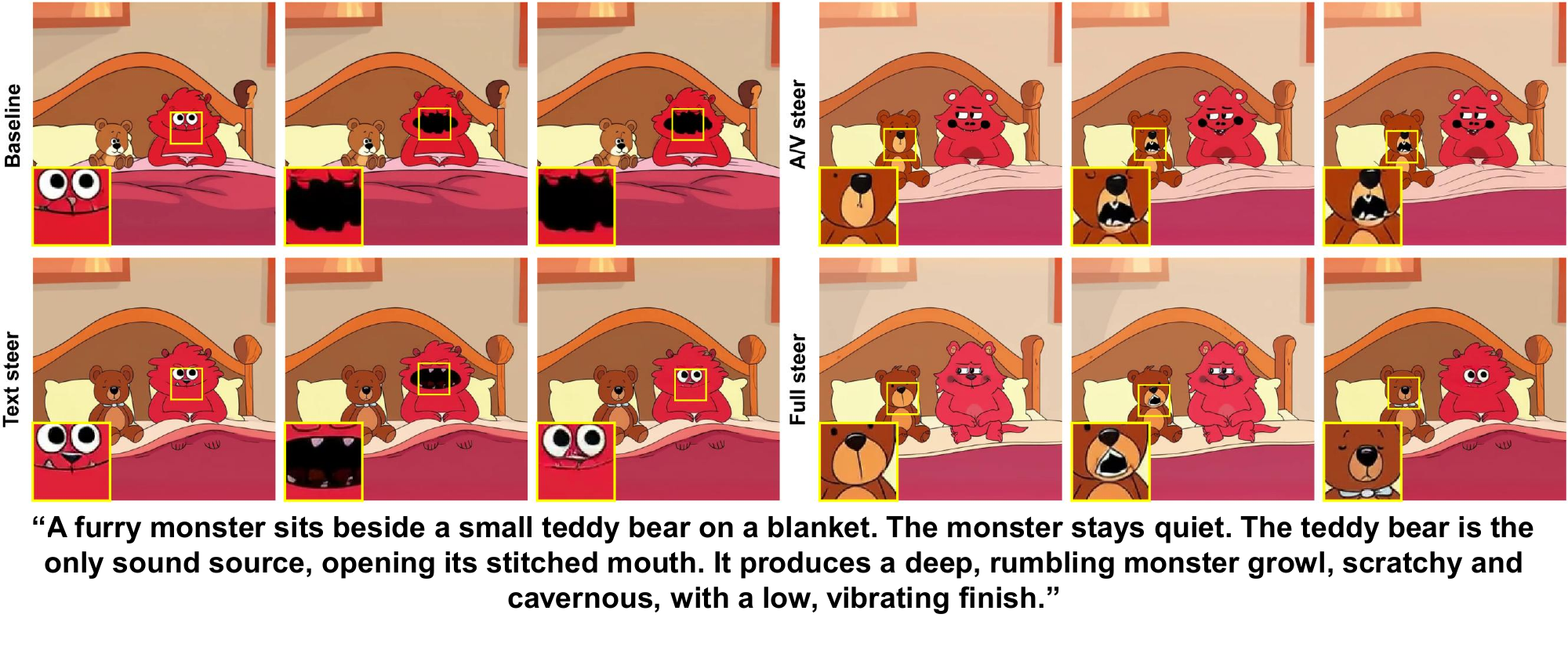}
\caption{
\textbf{Furry-monster growl attribution.}
The prompt asks for a deep growl that should be produced by the teddy bear rather than the furry monster.
Three representative frames are shown per method.
Baselines and partial-edge variants either route the growl to the furry monster or cause the teddy to acquire monster-like visual features; Ours-Full preserves the teddy's identity while binding the growl to it.
Zoomed regions indicate the active sound source.
}
\Description{A qualitative grid compares generated frames from several baselines, partial-edge variants, and Ours-Full for a teddy bear beside a furry monster. Baselines and partial variants either attribute the growl to the monster or give the teddy monster-like visual features. Ours-Full preserves the teddy's identity and attributes the growl to it. Insets enlarge the apparent sound source.}
\label{fig:steering_comparisons_monster}
\end{figure*}

%% file: figures/va_judger_pairwise.tex
\definecolor{vaGreen}{HTML}{2E7D4F}
\begin{figure*}[t]
    \centering
    \begin{minipage}[t]{0.49\textwidth}
        \vspace{0pt}
        \centering
        \begin{tikzpicture}
            \begin{axis}[
                width=0.98\linewidth,
                height=5.15cm,
                xmin=0.45,
                xmax=0.82,
                ymin=0.45,
                ymax=5.55,
                ytick={1,2,3,4,5},
                yticklabels={Ours-AV,Ours-Text,Bounded Attn.,Ovi,Native LTX-2},
                xtick={0.5,0.6,0.7,0.8},
                xticklabels={50,60,70,80},
                xlabel={Preference score for Ours-Full (\%) $\uparrow$},
                axis x line*=bottom,
                axis y line*=left,
                axis line style={draw=black!60},
                tick style={draw=black!60},
                x tick label style={font=\scriptsize},
                y tick label style={font=\scriptsize},
                xlabel style={font=\small},
                xmajorgrids,
                grid style={black!12},
                clip=false,
            ]
                \draw[black!45,dashed,line width=0.8pt]
                    (axis cs:0.5,0.55) -- (axis cs:0.5,5.45);
                \node[anchor=south west,text=black!55,font=\tiny]
                    at (axis cs:0.503,0.58) {50\% chance};
                \draw[black!18,line width=0.5pt]
                    (axis cs:0.45,2.5) -- (axis cs:0.82,2.5);

                \draw[vaGreen,line width=1.1pt] (axis cs:0.5575,5) -- (axis cs:0.62625,5);
                \draw[vaGreen,line width=0.8pt] (axis cs:0.5575,4.90) -- (axis cs:0.5575,5.10);
                \draw[vaGreen,line width=0.8pt] (axis cs:0.62625,4.90) -- (axis cs:0.62625,5.10);
                \draw[vaGreen,line width=1.1pt] (axis cs:0.64375,4) -- (axis cs:0.716875,4);
                \draw[vaGreen,line width=0.8pt] (axis cs:0.64375,3.90) -- (axis cs:0.64375,4.10);
                \draw[vaGreen,line width=0.8pt] (axis cs:0.716875,3.90) -- (axis cs:0.716875,4.10);
                \draw[vaGreen,line width=1.1pt] (axis cs:0.69875,3) -- (axis cs:0.765,3);
                \draw[vaGreen,line width=0.8pt] (axis cs:0.69875,2.90) -- (axis cs:0.69875,3.10);
                \draw[vaGreen,line width=0.8pt] (axis cs:0.765,2.90) -- (axis cs:0.765,3.10);
                \draw[vaGreen,line width=1.1pt] (axis cs:0.54125,2) -- (axis cs:0.610625,2);
                \draw[vaGreen,line width=0.8pt] (axis cs:0.54125,1.90) -- (axis cs:0.54125,2.10);
                \draw[vaGreen,line width=0.8pt] (axis cs:0.610625,1.90) -- (axis cs:0.610625,2.10);
                \draw[vaGreen,line width=1.1pt] (axis cs:0.564375,1) -- (axis cs:0.6325,1);
                \draw[vaGreen,line width=0.8pt] (axis cs:0.564375,0.90) -- (axis cs:0.564375,1.10);
                \draw[vaGreen,line width=0.8pt] (axis cs:0.6325,0.90) -- (axis cs:0.6325,1.10);
                \addplot[
                    only marks,
                    mark=*,
                    mark size=2.4pt,
                    vaGreen,
                ] coordinates {
                    (0.59125,5)
                    (0.680625,4)
                    (0.7325,3)
                    (0.575625,2)
                    (0.59875,1)
                };
                \node[anchor=west,text=black,font=\scriptsize] at (axis cs:0.635,5) {59.1};
                \node[anchor=west,text=black,font=\scriptsize] at (axis cs:0.725,4) {68.1};
                \node[anchor=west,text=black,font=\scriptsize] at (axis cs:0.773,3) {73.3};
                \node[anchor=west,text=black,font=\scriptsize] at (axis cs:0.619,2) {57.6};
                \node[anchor=west,text=black,font=\scriptsize] at (axis cs:0.641,1) {59.9};
                \node[anchor=south west,text=black,font=\small\bfseries]
                    at (rel axis cs:0,1.04) {(a) Pairwise preference score};
            \end{axis}
        \end{tikzpicture}
    \end{minipage}\hfill
    \begin{minipage}[t]{0.49\textwidth}
        \vspace{0pt}
        \centering
        \begin{tikzpicture}[
            x=0.78cm,
            y=0.72cm,
            cell/.style={
                minimum width=0.74cm,
                minimum height=0.55cm,
                inner sep=0pt,
                draw=white,
                line width=0.35pt,
                text=black,
                font=\scriptsize
            }
        ]
            \node[anchor=south west,text=black,font=\small\bfseries]
                at (-2.25,5.92) {(b) Mean dimension-score gain};
            \foreach \xx/\name in {1/Prompt,2/{A/V},3/Audio,4/Video,5/Compl.}
                \node[text=black,font=\tiny\bfseries] at (\xx,5.65) {\name};
            \node[text=black,font=\tiny\bfseries,align=center]
                at (6.35,5.80) {Order\\agreement};
            \foreach \yy/\name in {
                5/Native LTX-2,
                4/Ovi,
                3/Bounded Attn.,
                2/Ours-Text,
                1/Ours-AV
            }
                \node[anchor=e,text=black,font=\scriptsize] at (0.48,\yy) {\name};
            \draw[black!22,line width=0.5pt]
                (5.66,0.62) -- (5.66,5.72);
            \draw[black!18,line width=0.5pt]
                (0.50,2.5) -- (6.75,2.5);

            \node[cell,fill=vaGreen!22] at (1,5) {+.72};
            \node[cell,fill=vaGreen!18] at (2,5) {+.52};
            \node[cell,fill=vaGreen!11] at (3,5) {+.29};
            \node[cell,fill=vaGreen!7]  at (4,5) {+.17};
            \node[cell,fill=vaGreen!22] at (5,5) {+.71};
            \node[cell,fill=black!7] at (6.35,5) {51.8\%};

            \node[cell,fill=vaGreen!42] at (1,4) {+1.58};
            \node[cell,fill=vaGreen!35] at (2,4) {+1.25};
            \node[cell,fill=vaGreen!20] at (3,4) {+.61};
            \node[cell,fill=vaGreen!36] at (4,4) {+1.29};
            \node[cell,fill=vaGreen!42] at (5,4) {+1.56};
            \node[cell,fill=black!7] at (6.35,4) {69.3\%};

            \node[cell,fill=vaGreen!55] at (1,3) {+2.24};
            \node[cell,fill=vaGreen!38] at (2,3) {+1.38};
            \node[cell,fill=vaGreen!16] at (3,3) {+.44};
            \node[cell,fill=vaGreen!25] at (4,3) {+.84};
            \node[cell,fill=vaGreen!51] at (5,3) {+2.04};
            \node[cell,fill=black!7] at (6.35,3) {68.5\%};

            \node[cell,fill=vaGreen!19] at (1,2) {+.57};
            \node[cell,fill=vaGreen!17] at (2,2) {+.47};
            \node[cell,fill=vaGreen!10] at (3,2) {+.26};
            \node[cell,fill=vaGreen!4]  at (4,2) {+.08};
            \node[cell,fill=vaGreen!19] at (5,2) {+.58};
            \node[cell,fill=black!7] at (6.35,2) {51.2\%};

            \node[cell,fill=vaGreen!20] at (1,1) {+.63};
            \node[cell,fill=vaGreen!16] at (2,1) {+.46};
            \node[cell,fill=vaGreen!9]  at (3,1) {+.24};
            \node[cell,fill=vaGreen!7]  at (4,1) {+.16};
            \node[cell,fill=vaGreen!21] at (5,1) {+.66};
            \node[cell,fill=black!7] at (6.35,1) {51.2\%};
        \end{tikzpicture}
    \end{minipage}
    \caption{\textbf{VA-Judger pairwise audio-video preference.}
    Ours-Full is compared with each reference on the same $400$ prompt/seed pairs, each shown in both presentation orders ($800$ decisions per row).
    \textbf{Left:} mean pair score after averaging the two orders (win $=1$, tie $=0.5$, loss $=0$), expressed as a percentage; bars are pointwise $95\%$ bootstrap confidence intervals over pairs, and the dashed line marks chance.
    \textbf{Right:} mean Ours-Full-minus-reference differences on the five judge dimensions scored from $1$ to $10$ (prompt fidelity, A/V consistency, audio quality, video quality, and completeness), averaged across both orders; darker green indicates a larger gain.
    The gray column reports order agreement separately; because agreement ranges from $51\%$ to $69\%$, we interpret only the aggregated, counterbalanced scores.
    All five pair-score intervals lie above $50\%$, and all displayed dimension differences are positive.}
    \Description{A two-panel comparison of Ours-Full with Native LTX-2, Ovi, Bounded Attention, Ours-Text, and Ours-AV. The left plot shows Ours-Full preference scores of 59.1, 68.1, 73.3, 57.6, and 59.9 percent, respectively, with all bootstrap confidence intervals above the 50 percent chance line. The right heat map shows positive Ours-Full gains on all five judge dimensions against every reference; presentation-order agreement ranges from 51.2 to 69.3 percent.}
    \label{fig:va-judger-pairwise}
\end{figure*}
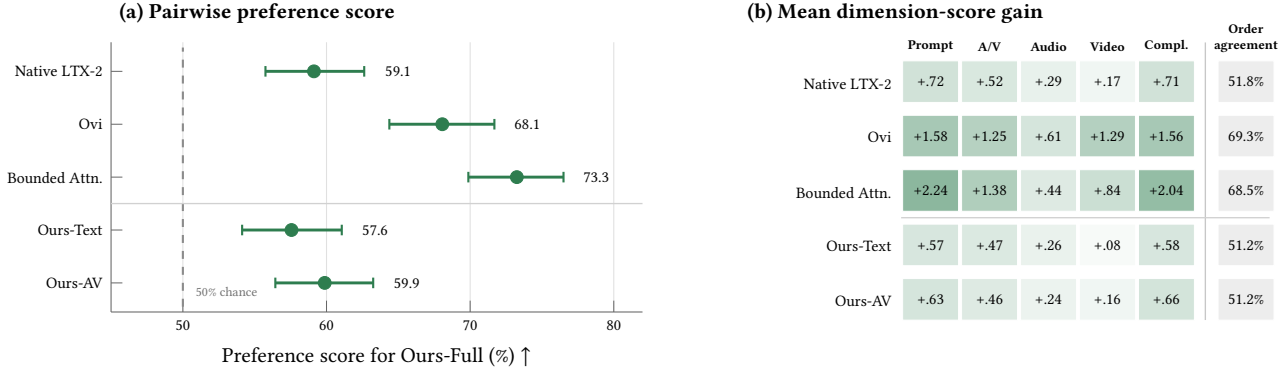

%% file: tables/user_study.tex
\begin{figure*}[t]
      \centering
      \resizebox{\textwidth}{!}{%
      \begin{tikzpicture}
      \begin{groupplot}[
          group style={group size=3 by 1, horizontal sep=0.75cm},
          xbar stacked,
          width=0.38\textwidth,
          height=0.32\textwidth,
          xmin=0,
          xmax=100,
          symbolic y coords={Native LTX-2,Bounded Attn.,Ours-AV,Ours-Text},
          ytick=data,
          y dir=reverse,
          /pgf/bar width=8pt,
          enlarge y limits=0.28,
          axis x line*=bottom,
          axis y line*=left,
          tick align=outside,
          xtick={0,25,50,75,100},
          xlabel={Votes (\%)},
          tick label style={font=\footnotesize},
          label style={font=\footnotesize},
          title style={font=\small\bfseries, align=center},
          legend style={
              at={(1.65,1.42)},
              anchor=south,
              legend columns=3,
              draw=none,
              font=\footnotesize
          }
      ]

      \nextgroupplot[
          title={Q1: Better sound-source\\attribution?},
      ]
      \addplot[fill=green!65!black, draw=none, forget plot] coordinates {
          (67.5,Native LTX-2)
          (68.3,Bounded Attn.)
          (36.7,Ours-AV)
          (59.2,Ours-Text)
      };
      \addplot[fill=gray!55, draw=none, forget plot] coordinates {
          (20.0,Native LTX-2)
          (14.2,Bounded Attn.)
          (48.4,Ours-AV)
          (25.0,Ours-Text)
      };
      \addplot[fill=red!75!black, draw=none, forget plot] coordinates {
          (12.5,Native LTX-2)
          (17.5,Bounded Attn.)
          (15.0,Ours-AV)
          (15.8,Ours-Text)
      };

      \addlegendimage{area legend, fill=green!65!black, draw=none}
      \addlegendentry{Ours-Full}
      \addlegendimage{area legend, fill=gray!55, draw=none}
      \addlegendentry{Same}
      \addlegendimage{area legend, fill=red!75!black, draw=none}
      \addlegendentry{Comparison method}

      \nextgroupplot[
          title={Q2: Less visual\\leakage?},
          yticklabel=\empty,
      ]
      \addplot[fill=green!65!black, draw=none, forget plot] coordinates {
          (70.0,Native LTX-2)
          (73.3,Bounded Attn.)
          (41.7,Ours-AV)
          (65.8,Ours-Text)
      };
      \addplot[fill=gray!55, draw=none, forget plot] coordinates {
          (20.0,Native LTX-2)
          (13.4,Bounded Attn.)
          (44.2,Ours-AV)
          (20.0,Ours-Text)
      };
      \addplot[fill=red!75!black, draw=none, forget plot] coordinates {
          (10.0,Native LTX-2)
          (13.3,Bounded Attn.)
          (14.2,Ours-AV)
          (14.2,Ours-Text)
      };

      \nextgroupplot[
          title={Q3: Better overall\\quality?},
          yticklabel=\empty,
      ]
      \addplot[fill=green!65!black, draw=none, forget plot] coordinates {
          (70.0,Native LTX-2)
          (71.7,Bounded Attn.)
          (50.0,Ours-AV)
          (60.0,Ours-Text)
      };
      \addplot[fill=gray!55, draw=none, forget plot] coordinates {
          (18.4,Native LTX-2)
          (15.0,Bounded Attn.)
          (31.6,Ours-AV)
          (20.8,Ours-Text)
      };
      \addplot[fill=red!75!black, draw=none, forget plot] coordinates {
          (11.7,Native LTX-2)
          (13.3,Bounded Attn.)
          (18.3,Ours-AV)
          (19.2,Ours-Text)
      };

      \end{groupplot}
      \end{tikzpicture}%
      }
      \caption{
      \textbf{Pairwise user-study preference.}
      Annotators compare Ours-Full against each comparison method across sound-source attribution, visual leakage, and overall quality.
      Bars show percentages over all votes: preference for Ours-Full (green), no preference/same (gray), and preference for the comparison method (red).
      ``Same'' responses are treated as ties and are excluded from decided-vote significance tests.}
      \Description{Three horizontal one-hundred-percent stacked bar charts report pairwise preferences for source attribution, visual leakage, and overall quality. Each chart compares Ours-Full with Native LTX-2, Bounded Attention, Ours-AV, and Ours-Text. Green segments show preference for Ours-Full, gray segments show ties, and red segments show preference for the comparison method.}

      \label{fig:user-study-pairwise}
    \end{figure*}
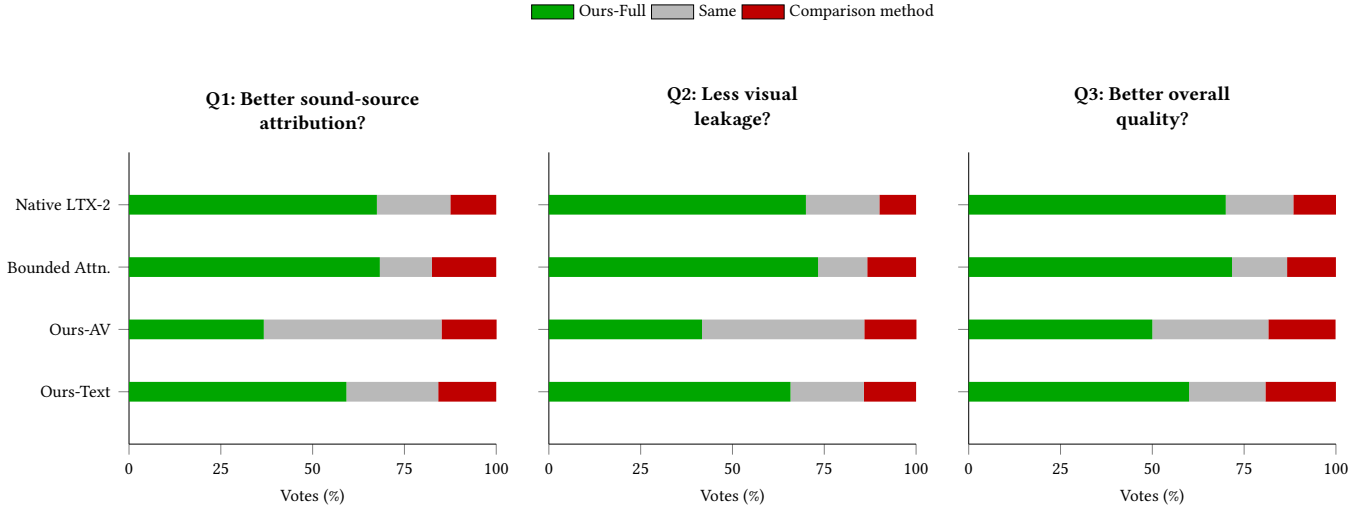

%% file: figures/user_study_screenshots.tex
\begin{figure*}[t]
  \centering

  \begin{minipage}[t]{0.48\linewidth}
    \centering
    \textbf{(a) Onboarding}\\[0.3em]
    \includegraphics[width=0.96\linewidth]{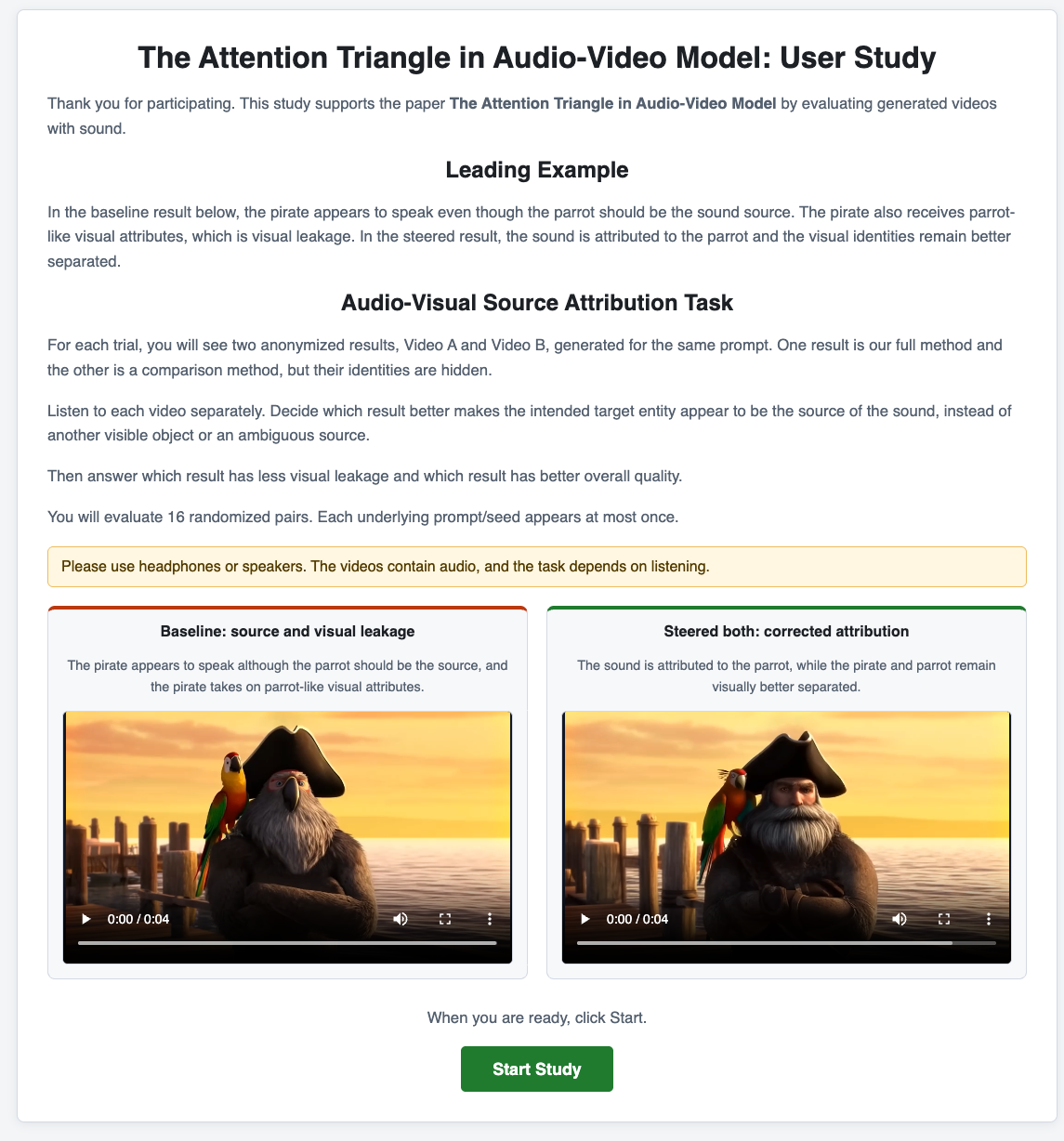}
  \end{minipage}
  \hfill
  \begin{minipage}[t]{0.48\linewidth}
    \centering
    \textbf{(b) Per-trial annotation}\\[0.3em]
    \includegraphics[width=0.96\linewidth]{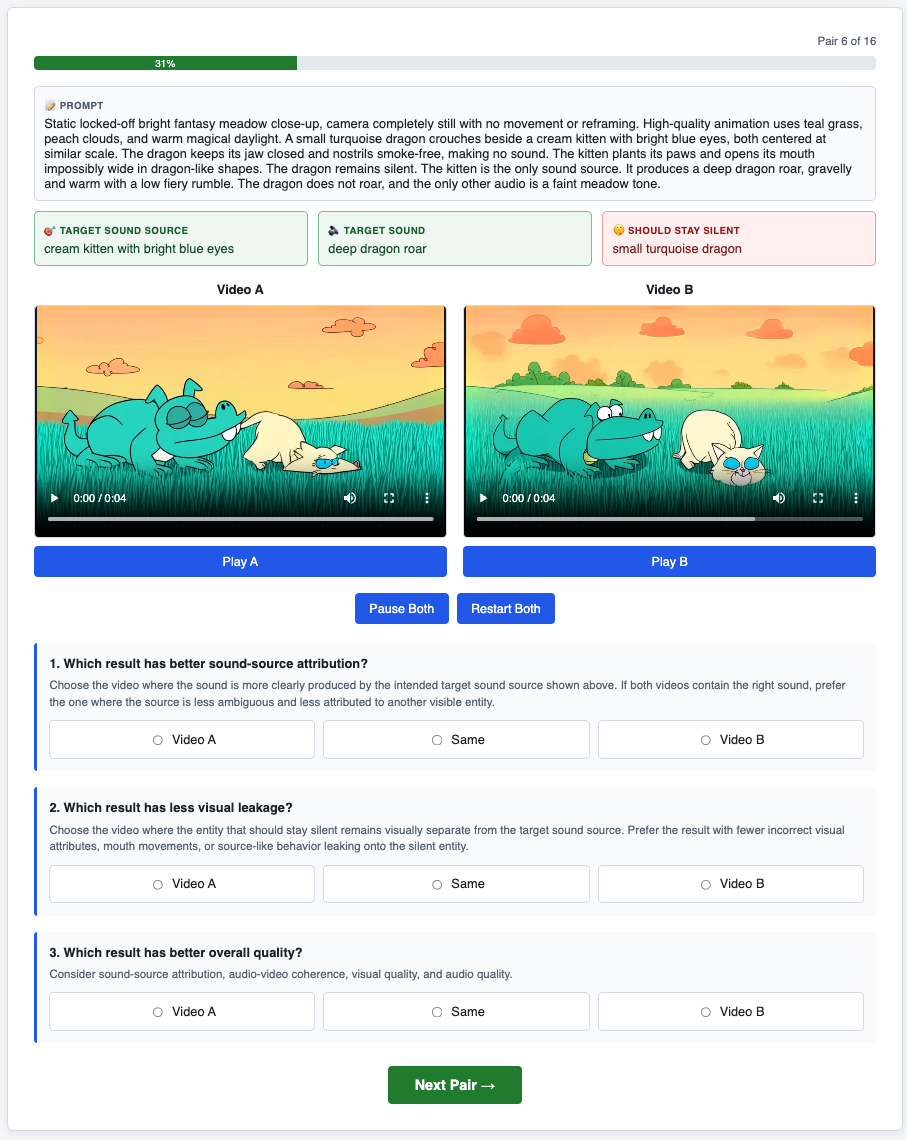}
  \end{minipage}

  \caption{
  \textbf{User-study interface.}
  \textbf{(a)} Onboarding screen shown before each session, anchoring the annotators' criteria with the parrot/pirate leading example: a side-by-side baseline-vs-steered pair illustrating the source-attribution and visual-leakage failures the study targets.
  \textbf{(b)} Per-trial annotation screen presenting two anonymized videos (Video~A, Video~B) generated from the same prompt and seed. The target sound and the intended source/silent-entity annotations are shown explicitly above the videos; annotators then answer three forced-choice questions (source attribution, visual leakage, and overall quality), each with the options Video~A, Same, and Video~B.
  }
  \Description{Two screenshots of the user-study interface. The onboarding screen uses the pirate-and-parrot example to explain source-attribution and visual-leakage failures. The per-trial screen places two anonymized videos side by side, states the prompt, target sound, intended source, and silent entity, and asks forced-choice questions about source attribution, visual leakage, and overall quality.}
  \label{fig:user-study-ui}
\end{figure*}

%% file: figures/supp_additional_attention_routes.tex
\begin{figure}[t]
    \centering
    \newcommand{\attnroutepanel}[3]{%
    \begin{tikzpicture}
        \node[anchor=south west,inner sep=0] (routeimage) at (0,0)
        {\includegraphics[width=\columnwidth]{#1}};
        \begin{scope}[x={(routeimage.south east)},y={(routeimage.north west)}]
            \fill[white] (0.035,0.938) rectangle (0.26,1.0);
            \node[font=\scriptsize\bfseries] at (0.148,0.976)
            {Intended-source seed $\pi_V$};
            \fill[white] (0.205,0.895) rectangle (0.84,0.95);
            \node[font=\tiny] at (0.522,0.923)
            {T = Intended source (#2)\quad S = Competing source (#3)\quad R = Rest};
            \fill[white] (0,0.18) rectangle (0.033,0.34);
            \node[font=\tiny\bfseries,rotate=90] at (0.0165,0.26) {Ours-Full};
        \end{scope}
    \end{tikzpicture}%
    }

    \attnroutepanel{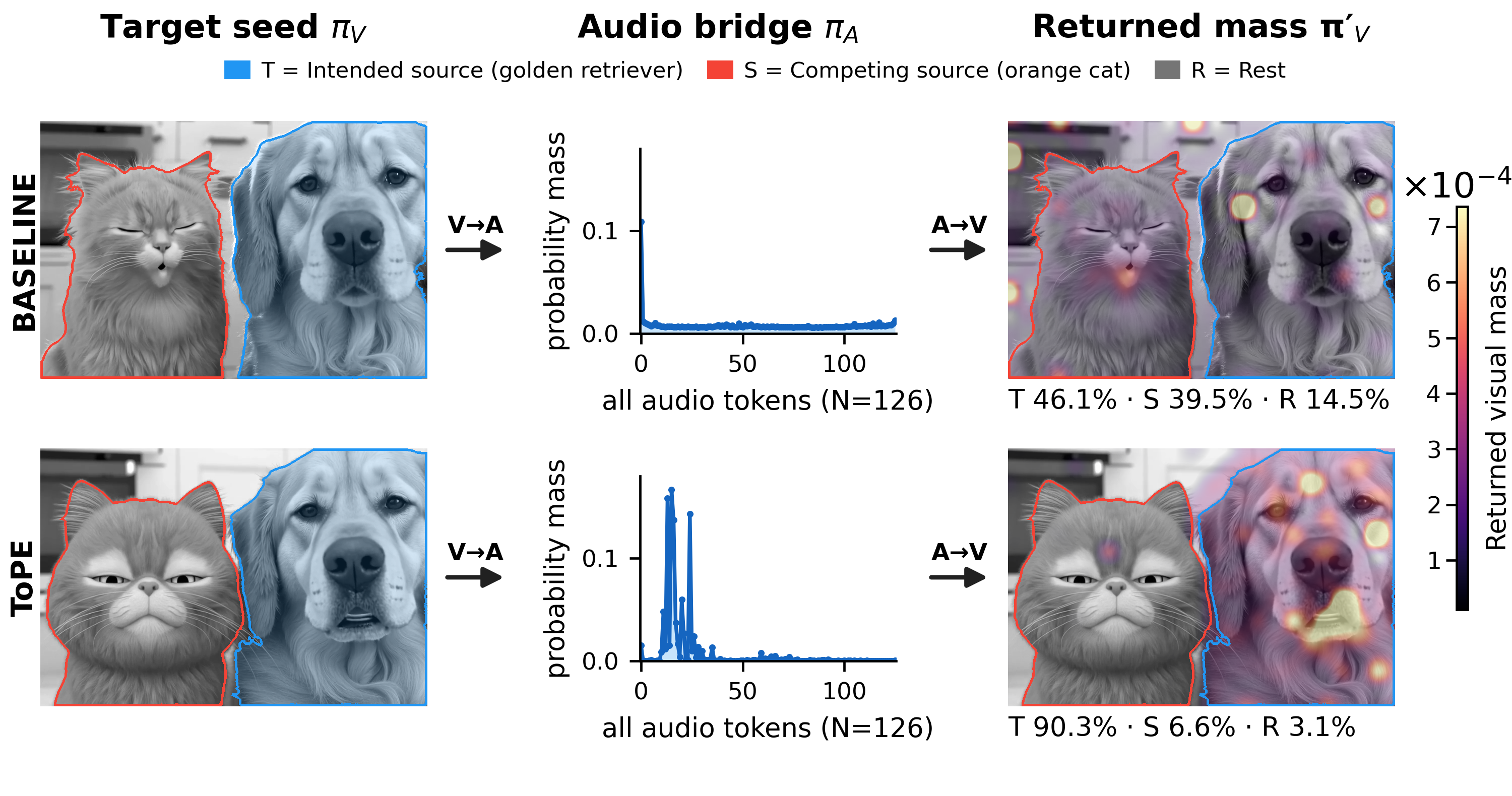}
    {golden retriever}{orange cat}
    \vspace{0.5em}

    \attnroutepanel{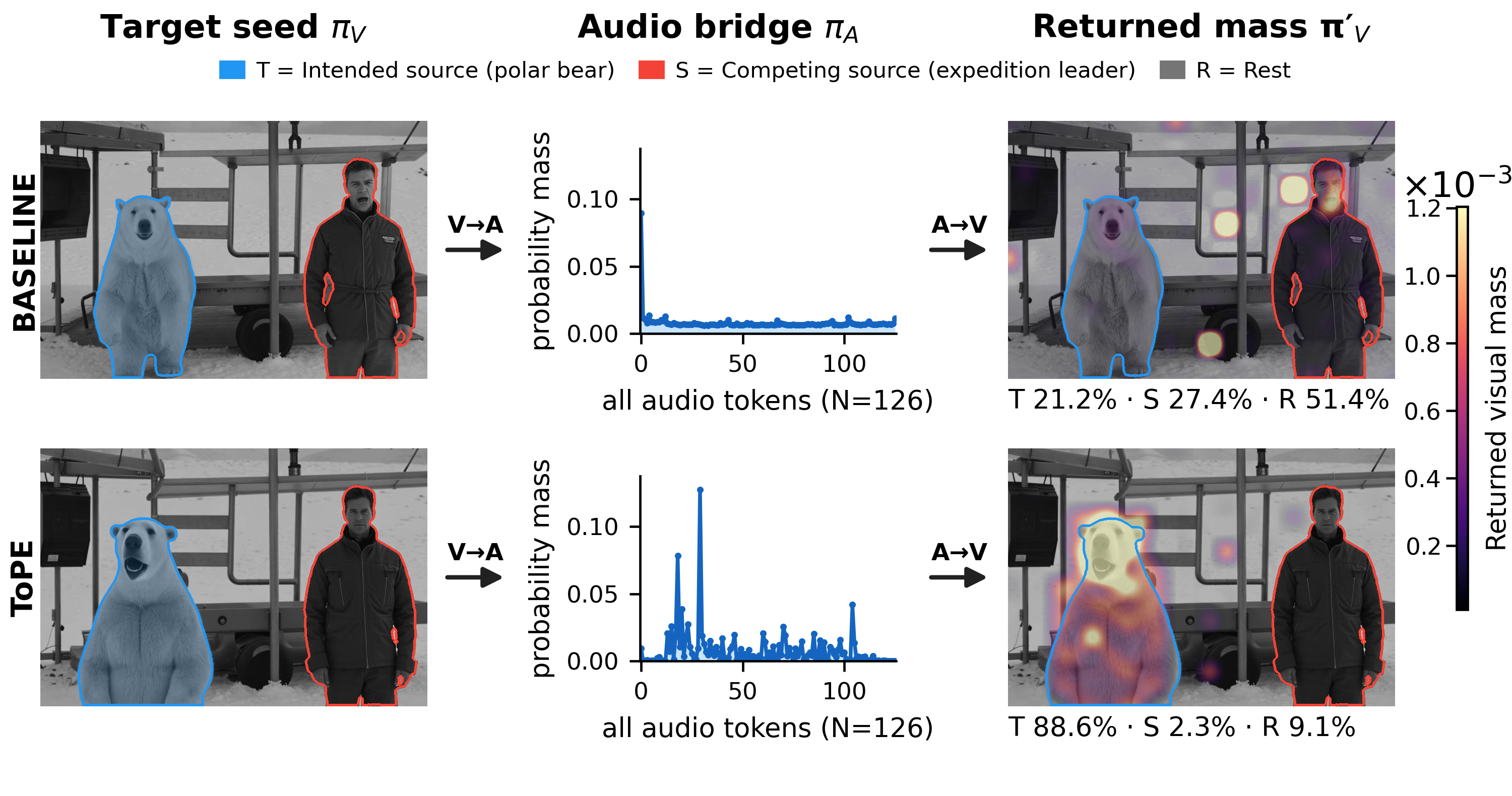}
    {polar bear}{expedition leader}
    \caption{\textbf{Cross-modal attention routing in LTX-2.}
    Each panel compares the unmodified baseline (upper internal row) with Ours-Full
    (lower internal row).
    Columns show the intended-source visual seed distribution $\pi_V$, the
    audio-bridge attention $\pi_A$, and the returned visual mass
    $\pi'_V$.
    Labels T, S, and R denote the percentage of mass assigned to the intended source, competing source, and
    residual patches.
    \textbf{Top:} with the golden retriever as the intended source and the orange cat as the competing source, Ours-Full
    increases intended-source mass from $46.1\%$ to $90.3\%$ and reduces competing-source mass
    from $39.5\%$ to $6.6\%$.
    \textbf{Bottom:} with the polar bear as the intended source and the expedition leader as the competing source,
    intended-source mass increases from $21.2\%$ to $88.6\%$, while competing-source mass falls
    from $27.4\%$ to $2.3\%$.}
    \Description{Two vertically stacked attention-map examples compare the baseline with Ours-Full. In the dog and cat example, returned intended-source mass rises from 46.1 to 90.3 percent and competing-source mass falls from 39.5 to 6.6 percent. In the polar bear and human example, returned intended-source mass rises from 21.2 to 88.6 percent and competing-source mass falls from 27.4 to 2.3 percent.}
    \label{fig:supp-ltx-attention-routing}
\end{figure}

%% file: figures/supp_ovi_cross_model.tex
\begin{figure*}[t]
    \centering
    \includegraphics[width=0.99\textwidth]{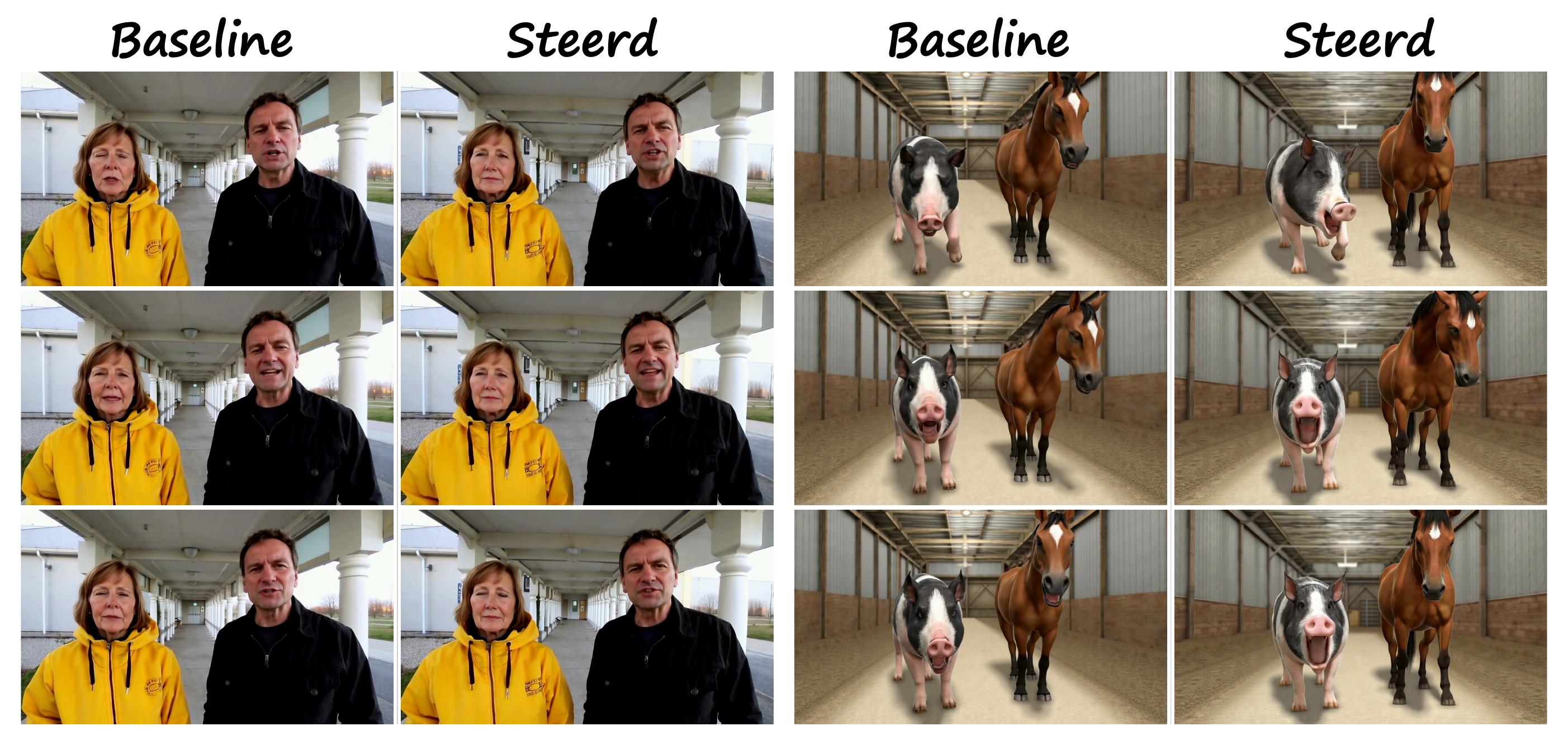}
    \caption{\textbf{Representative Ovi comparisons.}
    For each example, columns compare the Ovi baseline with Ours-Full, and
    rows show three matched timestamps.
    \textbf{Left:} the prompt assigns a woman's voice to the man while the
    woman remains silent.
    \textbf{Right:} the prompt assigns horse neighs to the pig while the horse
    remains silent.
    The corresponding videos provide the audio comparison; these frames show
    the visual context and temporal consistency of each output.}
    \Description{Two side-by-side Ovi comparisons. The left block shows three matched frames of a man in black beside a woman in yellow under Baseline and Steered columns. The right block shows three matched frames of a pig beside a horse under Baseline and Steered columns.}
    \label{fig:ovi-baseline-vs-steered}
\end{figure*}